\documentclass[10pt,journal]{IEEEtran}

\usepackage[utf8]{inputenc} 
\usepackage[T1]{fontenc}    
\usepackage{url}            
\usepackage{booktabs}       
\usepackage{amsfonts}       
\usepackage{nicefrac}       
\usepackage{microtype}      
\usepackage{graphicx}
\usepackage{amsmath}
\usepackage{amssymb}
\usepackage{makecell}
\usepackage{engord}

\usepackage{amsthm}
\newtheorem{theorem}{Theorem}[section]

\newtheorem{lemma}[theorem]{Lemma}

\newcommand{\R}[1]{\textcolor[rgb]{1.00,0.00,0.00}{#1}}
\newcommand{\B}[1]{\textcolor[rgb]{0.00,0.00,1.00}{#1}}

\usepackage{transparent}
\usepackage{bm}
\usepackage{multirow}
\usepackage[justification=centering]{subcaption}
\usepackage{nicefrac}
\usepackage{comment}
\usepackage{enumitem}
\usepackage{adjustbox}
\usepackage{tabu}
\usepackage{booktabs}
\usepackage{diagbox}
\usepackage{cuted}
\usepackage{capt-of}
\usepackage[table,x11names]{xcolor}
\usepackage[nointegrals]{wasysym} 
\usepackage{amsmath}
\usepackage{colortbl}
\usepackage{arydshln}

\usepackage{dblfloatfix}
\usepackage{transparent}

\usepackage{algorithm}
\usepackage{listings}

\usepackage{etoolbox}
\makeatletter
\AfterEndEnvironment{algorithm}{\let\@algcomment\relax}
\AtEndEnvironment{algorithm}{\kern0pt\hrule\relax\vskip0pt\@algcomment}
\let\@algcomment\relax
\newcommand\algcomment[1]{\def\@algcomment{\footnotesize#1}}
\renewcommand\fs@ruled{\def\@fs@cfont{\bfseries}\let\@fs@capt\floatc@ruled
\def\@fs@pre{\hrule height.8pt depth0pt \kern2pt}%
\def\@fs@post{}%
\def\@fs@mid{\kern2pt\hrule\kern2pt}%
\let\@fs@iftopcapt\iftrue}
\makeatother

\newcommand{\app}{\raise.17ex\hbox{$\scriptstyle\sim$}}

\newcolumntype{x}[1]{>{\centering\arraybackslash}p{#1pt}}
\newcolumntype{y}[1]{>{\raggedright\arraybackslash}p{#1pt}}
\newcolumntype{z}[1]{>{\raggedleft\arraybackslash}p{#1pt}}
\newlength\savewidth

\makeatletter\renewcommand\paragraph{\@startsection{paragraph}{4}{\z@}
{.5em \@plus1ex \@minus.2ex}{-.5em}{\normalfont\normalsize\bfseries}}\makeatother

\usepackage{textpos} 

\ifCLASSOPTIONcompsoc
  \usepackage[nocompress]{cite}
\else
  \usepackage{cite}
\fi

\ifCLASSINFOpdf
\else
\fi
\usepackage[pagebackref=true,breaklinks=true,letterpaper=true,colorlinks,bookmarks=false]{hyperref}

\usepackage[capitalize]{cleveref}
\crefname{section}{Sec.}{Secs.}
\Crefname{section}{Section}{Sections}
\Crefname{table}{Table}{Tables}
\crefname{table}{Tab.}{Tabs.}

\begin{document}

\title{Revisiting $\ell_1$ Loss in Super-Resolution: A Probabilistic View and Beyond}


\author{

Xiangyu He, Jian Cheng

\IEEEcompsocitemizethanks{

\IEEEcompsocthanksitem X. He and J. Cheng are with the 
National Laboratory of Pattern Recognition (NLPR), 
Institute of Automation, Chinese Academy of Sciences (CASIA), Beijing 100190, China
(e-mail: xiangyu.he@nlpr.ia.ac.cn; jcheng@nlpr.ia.ac.cn).
}
\thanks{Manuscript received April 19, 2005; revised August 26, 2015.}
}

\markboth{IEEE Transactions on Image Processing,~Vol.~XX, No.~XX, 2015}%
{He \MakeLowercase{\textit{et al.}}: Revisiting $\ell_1$ Loss in Super-Resolution: A Probabilistic View and Beyond}

\IEEEtitleabstractindextext{%
\inputsection{abstract}
}

\maketitle


\IEEEpeerreviewmaketitle

\begin{abstract}
Super-resolution as an ill-posed problem has many high-resolution candidates for a low-resolution input. However, the popular $\ell_1$ loss used to best fit the given high-resolution image fails to consider this fundamental property of non-uniqueness in image restoration. In this work, we fix the missing piece in $\ell_1$ loss by formulating super-resolution with neural networks as a probabilistic model. It shows that $\ell_1$ loss is equivalent to a degraded likelihood function that removes the randomness from the learning process. By introducing a data-adaptive random variable, we present a new objective function that aims at minimizing the expectation of the reconstruction error over all plausible solutions. The experimental results show consistent improvements on mainstream architectures, with no extra parameter or computing cost at inference time.
\end{abstract}


\IEEEpeerreviewmaketitle

\section{Introduction}
The behavior of single image super-resolution (SISR) networks is primarily driven by the choice of the objective function. Two common examples in SISR are the mean squared error (MSE) loss and the mean absolute error (MAE) loss. Since the long-tested measure peak signal-to-noise ratio (PSNR) is essentially defined on the MSE, pioneer works \cite{SRCNN, DRCN, FSRCNN, ESPCN, DRRN, SRGAN} have largely focused on minimizing the mean squared reconstruction error to achieve high performances.

Due to the ill-posed nature of SISR problem, a low-resolution image $\mathbf{x}$ can be well described as the downsampling result of many high-resolution images $\mathbf{\hat{y}}_{1:k}$. This implies that, in the inverse image reconstruction, the neural network trained by MSE loss, \textit{i.e.,} $\mathbb{E}_{(\mathbf{x},\mathbf{\hat{y}})}[\sum_i\Vert f(\mathbf{x})-\mathbf{\hat{y}}_i\Vert_2^2]$, will learn to output the \textit{mean} value of all HR targets, \textit{i.e.,} $f(\mathbf{x})\rightarrow\frac{1}{k}\sum_i\mathbf{\hat{y}}_i$, which best fits the image pairs $\{\mathbf{x},\mathbf{\hat{y}}_{1:k}\}$ \cite{MillerGS93}. However, the average of different HR images results in the absense of distinct edges and the blurring effect, which suffers from low image fidelity despite relatively high PSNR. Recently, $\ell_1$ loss has received much attention from low-level vision researches, since it is in practice more robust to the regression-to-the-mean problem in MSE loss \cite{Losses, EDSR}. Despite its emprical success in state-of-the-arts super-resolution methods \cite{RCAN, SAN, HAN, MeiFZHHS20, RFANet, IGNN}, $\ell_1$ loss itself has been rarely discussed in the context of image processing.

\begin{figure}[t]
\centering
\includegraphics[width=0.5\textwidth]{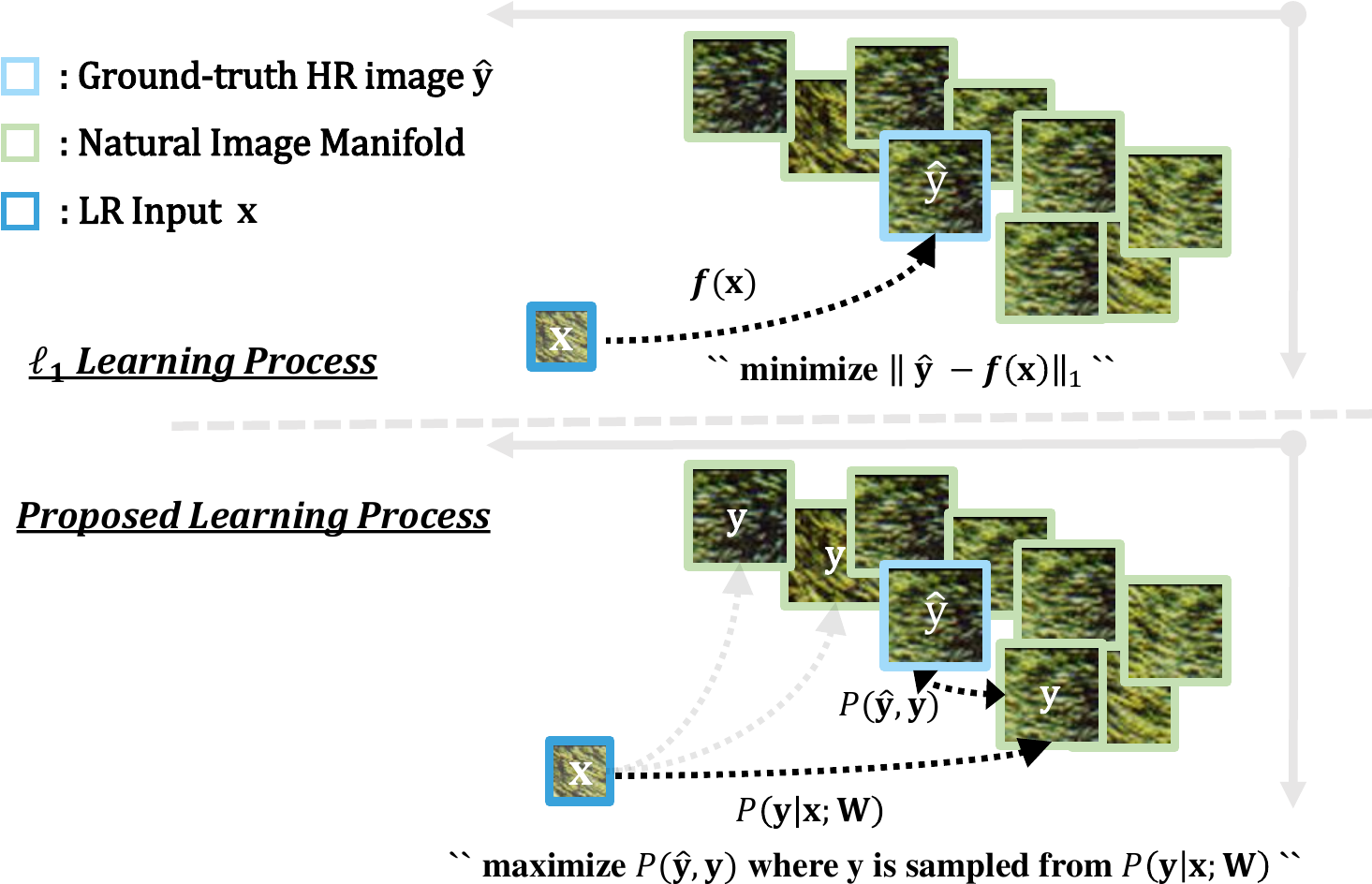}
\caption{Illustration of the learning process given an LR-HR image pair. $\ell_1$ loss only considers the fitting error between model output $f(\mathbf{x})$ and HR image $\mathbf{\hat{y}}$. We propose to learn an one-to-many mapping by introducing $P(\mathbf{y}|\mathbf{x};\mathbf{W})$ explicitly: 1) first perform a random sampling from $P(\mathbf{y}|\mathbf{x};\mathbf{W})$ to obtain a underlying natural image $\mathbf{y}$; 2) then maximize the likelihood of $\mathbf{y,\hat{y}}$ co-occurrence. Since we focus on the expected value averaged across all $\mathbf{y}$ instead of any specific sampling result, the learning process serves as a multiple-valued mapping.}
\label{motivation}
\end{figure}

In probabilistic setting, it is well known that a median is the minimizer of $\ell_1$ loss with respect to given targets \cite{stroock2010probability}. In the case of training neural networks over LR-HR image pairs, the situation is quite similar to MSE loss where the network will learn to recover the \textit{median} of all perceptually convincing HR images $\mathbf{\hat{y}}_{1:k}$. However, both $\ell_1$ and $\ell_2$ loss only capture the median/mean value of the whole solution space, which can be inadequate for estimating any specific solution in high-dimensional predictions, super-resolution in particular.

In this paper, we begin with a simple idea that \textit{it is intuitive to learn a one-to-many mapping since there are plenty of perceptually convincing solutions to an LR input} (shown in Figure \ref{motivation}). Ideally, we may approximate the local manifold, corresponding to the ground-truth HR image, by the posterior distribution $P(\mathbf{y}|\mathbf{x};\mathbf{W})$. Notice that, generative methods are able to capture the more complex distribution of natural images, which inspires us to revisit super-resolution with neural networks from a probabilistic view. In this way, we have the following contributions:
\begin{itemize}
\item We derive a loss function for solving SISR by introducing a posterior Gaussian distribution of underlying natural image $\mathbf{y}$, \textit{i.e.,} $P(\mathbf{y}|\mathbf{x};\mathbf{W})$, which minimizes the expected reconstruction error across all plausible solutions in $\mathbf{y}$. The formulated learning process can be easily extended to other $P(\mathbf{y}|\mathbf{x};\mathbf{W})$ with different distributions such as Laplace.
\item The learned standard deviation of $P(\mathbf{y}|\mathbf{x};\mathbf{W})$ facilitates the HR image regression and, as a by-product, allows us to make meaningful estimates of the model uncertainty in SISR, \textit{i.e.,} a way to reflect its SR correctness likelihood. This can be useful in settings where errors carry serious repercussions such as medical image super-resolution. 
\item The experimental results show consistent improvements over popular $\ell_1$ loss on mainstream networks, without the extra cost at runtime.
\end{itemize}

\section{Related Work}
\noindent\textbf{Super-resolution} Over the past decades, super-resolution is among the most fundamental low-level vision problems. Early works often formulate the task as an interpolation \cite{sampling, sampling2, sampling3, sampling4}. Latter studies start to reconstruct high-resolution images in a data-driven way, by estimating the natural image statistics \cite{prior, prior2} (including deep prior \cite{prior3, BrunaSL15}) and exploiting the neighbor embedding \cite{embedding, embedding2, embedding3}. Another group of methods utilizes the self-similarities to learn the inverse mapping \cite{self, self2, self3, self4}. Benefiting from the development of sparse coding, \cite{A, sparse, sparse3} achieve reasonable results on benchmark datasets.

Advances in deep learning have further enhanced the state-of-the-art performance since \cite{SRCNN, FSRCNN} first introduce a CNN-based SR method. Various CNN architectures have been proposed such as residual networks \cite{VDSR, DRCN, EDSR}, densely connected networks \cite{RDN, 0001LLG17}, ode-inspired design \cite{OISR} and U-Net encoder-decoder networks \cite{MaoSY16, ChengMLZZZ19, LiuZZLZ18}. In particular, the attention mechanism has become a popular tool for image super-resolution. Both channel attention \cite{RCAN} and non-local technique \cite{HAN, SAN} achieve notable improvements over baseline methods. Recently, self-correlation/self-attention that models the long-distance relationship among similar patches have led to successful results \cite{IGNN, MeiFZHHS20, NLSA}, as the large training dataset can be used as a means to capture the global information \cite{Transformer, swinIR, YangYFLG20}. Our approach can facilitate the training of deep learning-based methods in a plug-in and play manner.

\noindent\textbf{Loss functions} Most learning objects in the literature fall into two main categories: pixel-wise loss and perceptual loss. The former aims to optimize the full reference metric such as PSNR and SSIM. MSE loss, in particular, is equivalent to PSNR. However, \cite{DosovitskiyB16, JohnsonAF16, MathieuCL15} show that $\ell_2$ loss results in overly-smooth SR images and the following works tend to use $\ell_1$ during training. In light of the success of knowledge distillation in image recognition, \cite{PISR, GaoZL018} use loss functions in the feature space to enhance the performance of student networks. Our work also falls into this category. We propose to learn from a posterior distribution instead of best-fitting ground-truth images.

Perceptual loss tackles the problem by employing generative adversarial networks \cite{MathieuCL15, DentonCSF15}, especially in super-resolution with large upscaling factors (\textit{e.g.,} $4\times$, $8\times$) \cite{YuP16, SRGAN}. The ImageNet pre-trained networks are commonly used as a discriminator loss \cite{JohnsonAF16, DosovitskiyB16, BrunaSL15}, which leads to high-frequency details and perceptually satisfying results in the sense. Our approach can be compatible with those adversarial settings as the content loss term in a perceptual loss.

The most related work is Noise2noise \cite{Noise2Noise}, which suffers from inaccurate gradient directions (details in Section \ref{regularization}). We solve this problem by a well-bounded gradient step and point out that \cite{Noise2Noise} indeed serves as a special case of our approach. Noise2Self \cite{Noise2Self} assumes that noise in each patch is independent of other patches, which can not hold responsible when it comes to the content of images instead of noise. Self2Self \cite{Self2Self} presents dropout in both image and feature space to reduce prediction variances, which is also compatible with our loss functions.

\noindent\textbf{Model uncertainty} Despite their unprecedented power to solve the inverse problem, deep learning-based SR methods are prone to make mistakes. As deep models are widely used in computer-aided diagnosis \cite{YueSLYZZ16, ChenSCXZL18}, it is crucial to better understand the confidence of neural networks in their predictions. That is, the SR network may provide uncertainty about the generated super-resolved images: which part is ``real'' and when we should be cautious.

\cite{Ghahramani15} first develops a framework for modeling uncertainty, which is hard to describe the modern deep neural networks. Recently, Monte Carlo Dropout \cite{GalG16} presents a practical solution to estimate uncertainty via the dropout layer. \cite{TeyeAS18} further shows that networks trained with Batch-Normalization \cite{IoffeS15} (BN) is an approximate Bayesian model and allows an uncertainty estimation by sampling the network's BN parameters. Unfortunately, neither of them was used in recent SR networks, which disables the variational inference in those approximate Bayesian models. Instead, we model the posterior distribution $P(\mathbf{y}|\mathbf{x};\mathbf{W})$ explicitly, as a by-product, the estimated standard deviation facilitates the uncertainty estimation for SISR.

\section{Analysis}
\subsection{Probabilistic Modeling}
From a statistical viewpoint, we are interested in the point estimation of the model parameters $\mathbf{W}$, used in the following likelihood function:
\begin{align}
\max_\mathbf{W}\ \mathcal{L}(\mathbf{W}|\mathbf{\hat{y}})=P(\mathbf{\hat{y}}|\mathbf{x};\mathbf{W})
\label{ML}
\end{align}
where $\mathbf{W}$, $\mathbf{\hat{y}}$ and $\mathbf{x}$ indicate weights, observed ground-truth labels and inputs respectively. Super-resolution using convolutional neural networks also fall into this form, that learning parameters $\mathbf{W}$ w.r.t. an inference function $f(\mathbf{x})$ from LR-HR image pairs directly.

In contrast to finding a one-to-one mapping between corrupted images and targets implied by $f(\mathbf{x})$, we begin with a subtle point: in reality, the mapping from $\mathbf{\hat{y}}$ to $\mathbf{x}$ is surjective. For example, a single LR image $\mathbf{x}$ may correspond to the downsampling result of multiple natural images. Hence, \textit{it is intuitive to recover or approximate the local image manifold instead of simply fitting the high-resolution image $\mathbf{\hat{y}}$}.

\noindent\textbf{Objective function} The idea is to explicitly estimate a posterior distribution $P(\mathbf{y}|\mathbf{x};\mathbf{W})$ where $\mathbf{y}$ refers to (multiple) underlying natural images with the size of $H\times W$. Then, the observed HR image $\mathbf{\hat{y}}$ should look like its counterpart $\mathbf{y}$ sampled from $P(\mathbf{y}|\mathbf{x};\mathbf{W})$, with high probability. Formally, our goal is to maximize the following conditional joint density function:
\begin{equation}
P(\mathbf{\hat{y}},\mathbf{y}|\mathbf{x};\mathbf{W})=P(\mathbf{\hat{y}|y})P(\mathbf{y}|\mathbf{x};\mathbf{W})
\label{likelihood}
\end{equation}
where $P(\mathbf{\hat{y}|y})$ is a measure of the probability that $\mathbf{\hat{y}}$ being observed given $\mathbf{y}$. Note that, there are plenty of $\mathbf{y}$ can serve as the perceptually convincing solution to Equation (\ref{likelihood}) then, as is common in machine learning, we take the expectation of that formula to be optimized:
\begin{align}
\mathbb{E}_{\mathbf{y}\sim P(\mathbf{y}|\mathbf{x};\mathbf{W})}[P(\mathbf{\hat{y}|y})].
\label{expectation}
\end{align}
To solve Equation (\ref{expectation}), we will first need to deal with $P(\mathbf{y}|\mathbf{x};\mathbf{W})$, the distribution of underlying natural images $\mathbf{y}$ consistent with the low-resolution $\mathbf{x}$.

\noindent\textbf{Optimizing the objective} Based on the finding of \textit{non-local means} \cite{BuadesCM05, nonlocalmeans} that each pixel is obtained as a weighted average of pixels centered at regions similar to the reference and the central limit theorem (CLT), we shall model $P(\mathbf{y}|\mathbf{x};\mathbf{W})$ as a multivariate normal distribution\footnote[1]{The distribution of $P_{\mathbf{y|x}}$ is not required to be Gaussian: for instance, if $\mathbf{y}$ is in practice more-Laplace, then $P_{\mathbf{y|x}}$ might be re-parameterized as $\bm\mu-\bm\sigma*\mathrm{sgn}(\mathbf{z})*\mathrm{ln}(1-2|\mathbf{z}|)$, $\mathbf{z}\sim \mathcal{N}(\bm0,\bm1)$.}:
\begin{equation}
P(\mathbf{y}|\mathbf{x};\mathbf{W})\sim\mathcal{N}(\bm{\mu}_{(\mathbf{x};\mathbf{W})};\bm{\Sigma}_{(\mathbf{x};\mathbf{W})})
\end{equation}
where mean $\bm{\mu}$ and covariance $\bm{\Sigma}$ are outputs of multi-layer neural networks with parameters $\mathbf{W}$. $\bm\sigma^2$ corresponds to the diagonal elements in $\bm\Sigma$. However, the expectation in Equation (\ref{expectation}) still involves sampling $\mathbf{y}$ from $P(\mathbf{y}|\mathbf{x};\mathbf{W})$ which is a non-differentiable operation. To allow the backprop through $\mathbf{y}$, we use one of the most common trick in machine learning, the ``reparameterization'' \cite{VAE}, to move the sampling to the random variable $\mathbf{z}\sim \mathcal{N}(\bm0,\bm1)$:
\begin{equation*}
\mathbb{E}_{\mathbf{y}\sim P(\mathbf{y}|\mathbf{x};\mathbf{W})}[P(\mathbf{\hat{y}|y})]=\mathbb{E}_{\mathbf{z}\sim \mathcal{N}(\bm0,\bm1)}[P(\mathbf{\hat{y}|\bm\mu+\bm\sigma*z})]
\end{equation*}
where $\mathbf{y}$ is reformulated as $\bm\mu+\bm\sigma*\mathbf{z}$ and ``$*$'' indicates elementwise product. Then, we can conduct the standard gradient descent and the gradient averaged over $\mathbf{z}$ should converge to the actual gradient of Equation (\ref{expectation}).

Similarly, as \cite{BrunaSL15}, we consider the conditional model defined by the Gibbs distribution (also known as Boltzmann distribution):
\begin{equation}
P(\mathbf{\hat{y}|y})\varpropto\prod_i^{H\times W}\mathrm{exp}(-\frac{|\mathbf{\hat{y}}_i-\mathbf{y}_i|}{kT})
\label{cond}
\end{equation}
where the corresponding Gibbs energy is in the form of $\Vert\mathbf{\hat{y}}-\mathbf{y}\Vert_1$ and $kT$ is the fixed partition function. The model (\ref{cond}) assumes that $\mathbf{y}$, looks like the observed ground-truth image $\mathbf{\hat{y}}$, is likely to be a strong evidence for $\mathbf{\hat{y}}$ occurring. We can now get both $P(\mathbf{y}|\mathbf{x};\mathbf{W})$ and $P(\mathbf{\hat{y}|y})$ into Equation (\ref{expectation}) then apply negative log-likelihood to the final objective for computational convenience:
\begin{align}
\min\ \mathbb{E}_{\mathbf{z}}\Big[\frac{1}{kT}\sum_{i}^{H\times W}|\mathbf{\hat{y}}_i-(\bm\mu_i+\bm\sigma_i*\mathbf{z})|\Big].
\label{eq:loss}
\end{align}
This equation implies a two-branch network ($\bm\mu$ and $\bm\sigma$) that requires randomness during training. The expectation on the data-independent random variable $\mathbf{z}$ allows us to move the gradient symbol into the expectation to finish the standard stochastic gradient descent \cite{VAE_tutorial}. For notational simplicity, we shall in the remaining text denote Equation (\ref{eq:loss}) as $\mathbb{E}_\mathbf{z}\Big[\Vert\mathbf{\hat{y}}-(\bm\mu+\bm\sigma*\mathbf{z})\Vert_1\Big]$.

\subsection{Revisiting $\ell_1$ Loss}\label{sect:degrad}
According to Jensen's inequality, for any measurable convex function $f(\cdot)$, we have $\mathbb{E}[f(\mathbf{z})]\geq f(\mathbb{E}[\mathbf{z}])$. Since every $p$-norm is convex and $\mathbf{z}\sim \mathcal{N}(\bm0,\bm1)$, then we have
\begin{align}
\mathbb{E}_\mathbf{z}\Big[\Vert\mathbf{\hat{y}}-(\bm\mu+\bm\sigma*\mathbf{z})\Vert_1\Big]&\geq\Vert\mathbf{\hat{y}}-(\bm\mu+\bm\sigma*\mathbb{E}_\mathbf{z}[\mathbf{z}])\Vert_1
\end{align}
Considering $\mathbb{E}_\mathbf{z}[\mathbf{z}]=0$, then
\begin{align}
\mathbb{E}_\mathbf{z}\Big[\Vert\mathbf{\hat{y}}-(\bm\mu+\bm\sigma*\mathbf{z})\Vert_1\Big]\geq\Vert\mathbf{\hat{y}}-\bm\mu\Vert_1.
\end{align}
Notice that the proposed learning target (\ref{eq:loss}) actually serves as an upper bound of the popular $\ell_1$ loss function. In other words, $\ell_1$ loss also suffers from the degradation that loses the randomness during training and is confined to the given HR images $\mathbf{\hat{y}}$. Though the optimal solution for the original function and for the upper bound are indeed different, the probabilistic learning object can relax the deterministic learning process at each step by focusing on the expectation of the loss function.

\begin{figure}[t]
\begin{minipage}[t]{.155\textwidth}
\centering
\includegraphics[width=\textwidth]{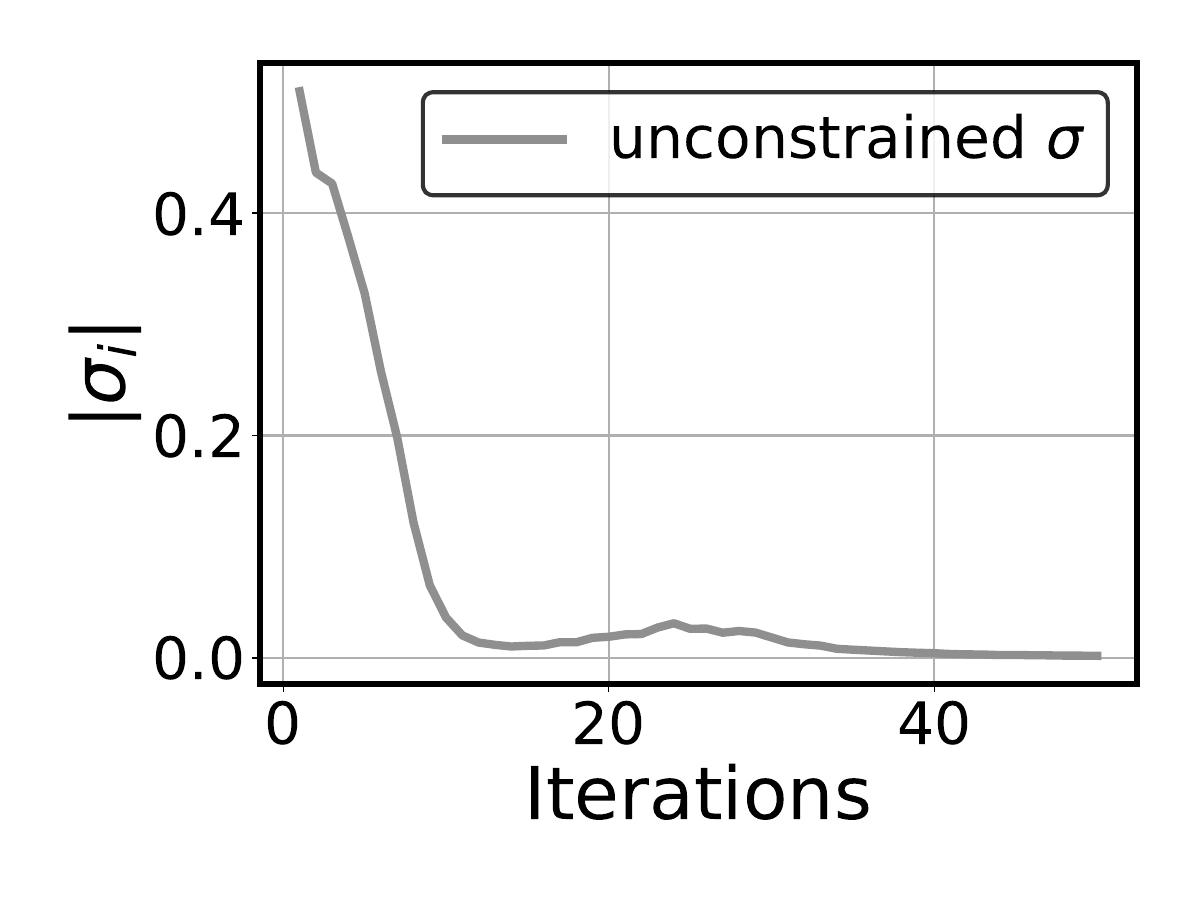}
\subcaption{R channel}
\end{minipage}
\begin{minipage}[t]{.155\textwidth}
\centering
\includegraphics[width=\textwidth]{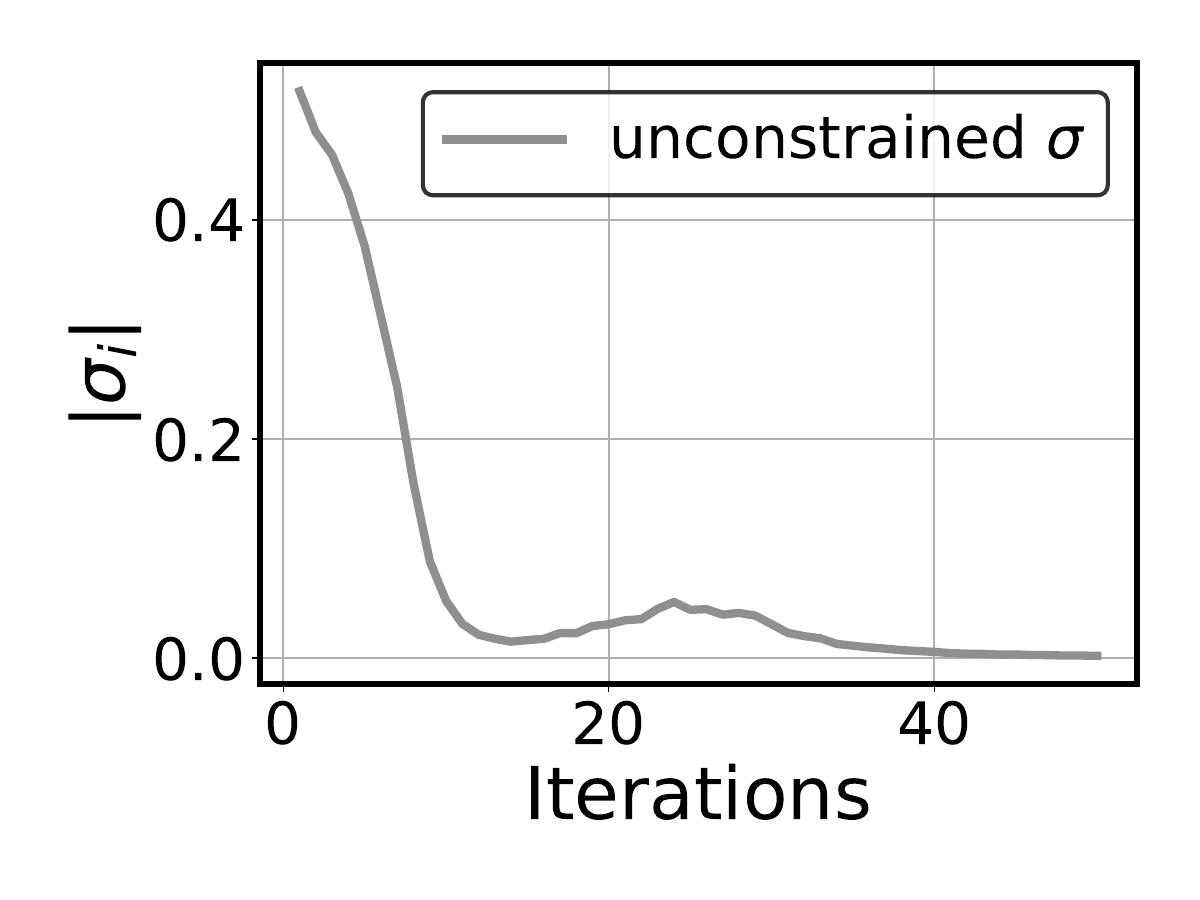}
\subcaption{G channel}
\end{minipage}
\begin{minipage}[t]{.155\textwidth}
\centering
\includegraphics[width=\textwidth]{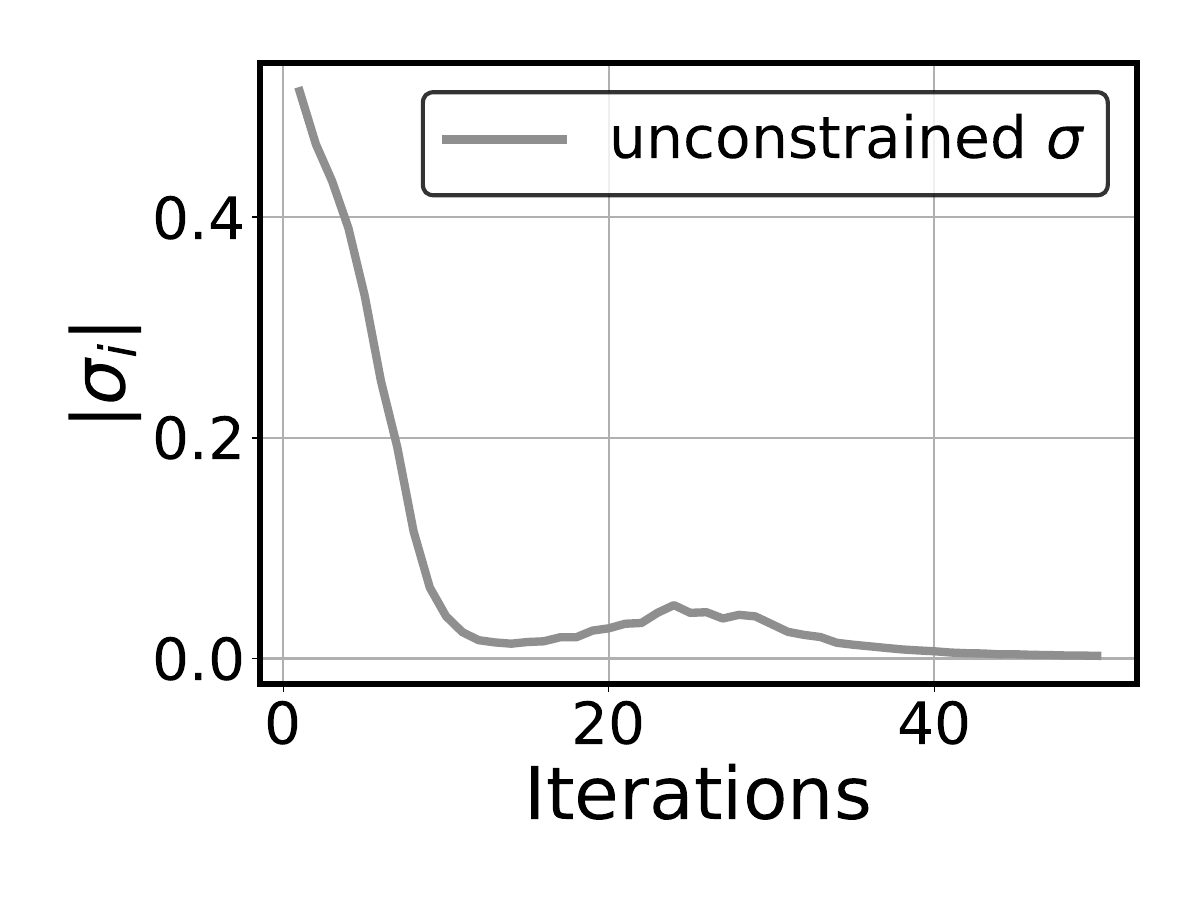}
\subcaption{B channel}
\end{minipage}
\caption{The degradation curves of $\bm\sigma_i$ during training. When $\bm\sigma_i$ approaching zero, Equation (\ref{eq:loss}) is equivalent to $\ell_1$ loss.}
\label{fig:sigma_decrease}
\end{figure}

Owing to the reparameterization trick \cite{VAE}, learning probabilistic model via Equation (\ref{eq:loss}) seems straightforward. However, there exists a trivial solution to easily bridge the Jensen gap (\textit{i.e.,} $\mathbb{E}[f(\mathbf{z})]=f(\mathbb{E}[\mathbf{z}])$) such that $\bm\sigma\in\vec{\mathbf{0}}$. As shown in Figure \ref{fig:sigma_decrease}, the output of the $\bm\sigma$ branch gradually approaches zero which ultimately eliminates the randomness in our learning target. Similar degradation also occurs in Variational Auto-Encoder (VAE). In a nutshell, VAE is an auto-encoder when latent variable $\mathbf{z}$ that determines encodings distribution lose randomness. VAE regularises its random latent variable $\mathbf{z}$ towards Normal distribution that enables generative process \cite{VAE_tutorial}. Otherwise, the latent space will converge to a single point instead of a distribution. That is, VAE degrades into an one-to-one mapping given input images, \textit{i.e.,} an (deterministic) auto-encoder. In light of this, we focus on avoiding the degradation of $\bm\sigma$ in the following section.

\subsection{Beyond $\ell_1$ Loss}\label{regularization}
\noindent\textbf{Data-independent $\bm\sigma$} The simplest way is to remove the trainable $\bm\sigma$ branch and let $\bm\sigma_i$ be a small constant $k$ invisible to the human eye. Then, $P(\mathbf{y}|\mathbf{x};\mathbf{W})\sim\mathcal{N}(\bm\mu;k^2\bm I)$ where $\bm I$ refers to an identity matrix. Contracting this term into Equation (\ref{eq:loss}) and rearranging yields:

\begin{equation}
\mathbb{E}_\mathbf{z}\Big[\Vert\mathbf{\hat{y}}-(\bm\mu+k\cdot\mathbf{z})\Vert_1\Big]= \mathbb{E}_\mathbf{z}\Big[\Vert\underbrace{\mathbf{(\hat{y}}+\mathbf{z'})}_{\mathrm{Noised\ HR}}-\bm\mu\Vert_1\Big]
\label{eq:noise2noise}
\end{equation}
where $\mathbf{z'}$ is an additive Gaussian noise $\mathcal{N}(0,k)$ and $k\neq0$. The left hand side of this equation is something we hope to optimize via backward propagation. The right hand side serves is the core of Noise2Noise \cite{Noise2Noise} that restores corrupted images without clean data, \textit{i.e.}, learning super-resolution by only looking at noised HR images.

Though the performance of training on corrupted observations may approach its clean counterpart \cite{Noise2Noise}, the noise term can do harm to the optimization process. To be specific, we expect $\bm\mu$ to approach $\mathbf{\hat{y}}$ via stochastic gradient descent. However, for some $\mathbf{z'}$ satisfying $(\mathbf{(\hat{y}}_i-\bm\mu_i)+\mathbf{z'})(\mathbf{\hat{y}}_i-\bm\mu_i)<0$, we have
\begin{align*}
\mathrm{sgn}(\frac{\partial \Vert\mathbf{(\hat{y}}+\mathbf{z'})-\bm\mu\Vert_1}{\partial\bm\mu_i})=-1\cdot\mathrm{sgn}(\frac{\partial \Vert\mathbf{\hat{y}}-\bm\mu\Vert_1}{\partial\bm\mu_i})
\end{align*}
where $\mathbf{z'}$ \textit{results in gradient ascent instead of descent, which ultimately hinders $\bm\mu$ from fitting $\mathbf{\hat{y}}$}. To get a good estimate of the proper gradient, it would require sampling massive noised observations given a single clean image, which would be expensive. Hence, we hope to find some $\bm\sigma$ that run correct backprop efficiently and retain randomness.

\noindent\textbf{Data-adaptive $\bm\sigma$} Fortunately, a delicate $\bm\sigma$ gives a definite answer to both. The full equation we want to optimize is:
\begin{align}
\min_{\bm\mu}\ \mathbb{E}_\mathbf{z}\Big[\big\Vert\mathbf{\hat{y}}-(\bm\mu+|\mathbf{\hat{y}}-\bm\mu|*\mathbf{z})\big\Vert_1\Big]
\label{main_loss}
\end{align}
where $|\mathbf{\hat{y}}-\bm\mu|$ corresponds to the $\bm\sigma$ term and $|\cdot|$ refers to element-wise absolute function. We shall denote Equation (\ref{main_loss}) as $\ell_\mathbb{E}$ in the remaining text for simplicity.

From the view of Equation (\ref{eq:noise2noise}), $\ell_\mathbb{E}$ is also in the form of adding some (data-adaptive) noise to HR images. However, \textit{for Equation (\ref{main_loss}), we allow the backprop through $|\mathbf{\hat{y}}-\bm\mu|$ instead of regarding it as a scaling factor}, which contributes to the following property: 

\begin{lemma}
Given a random variable $\mathbf{z}\sim\mathcal{N}(\bm0,\bm1)$ and fixed $\mathbf{\hat{y}}$, the gradient direction of $\bm\mu$ with respect to the following object
\begin{equation*}
\min_{\bm\mu}\ \mathbb{E}_\mathbf{z}\Big[\big\Vert(\bm\mu+|\bm\mu-\mathbf{\hat{y}}|*\mathbf{z})-\mathbf{\hat{y}}\big\Vert_1\Big]
\end{equation*}
is always consistent with $\Vert\bm\mu-\mathbf{\hat{y}}\Vert_1$ and $\Big|\mathbb{E}_\mathbf{z}\big[\frac{\partial \ell_\mathbb{E}}{\partial\bm\mu_i}\big]\Big|$ converges to $1.17$.
\end{lemma}

\begin{proof}
Since the expectation of random variable $\mathbf{z}$ does not depend on model parameters, we can safely move the gradient symbol into $\mathbb{E}_\mathbf{z}[\cdot]$ while maintaning equality:
\begin{equation*}
\frac{\partial\ell_\mathbb{E}}{\partial\bm\mu_i}=\mathrm{sgn}\big(\bm\mu_i+|\bm\mu_i-\mathbf{\hat{y}}_i|*\mathbf{z}_i-\mathbf{\hat{y}}_i\big)\big(1+\mathbf{z}_i*\mathrm{sgn}(\bm\mu_i-\mathbf{\hat{y}}_i)\big).
\end{equation*}
Assuming that $\bm\mu_i>\mathbf{\hat{y}}_i$, the above equation can be further simplified as:
\begin{equation*}
\frac{\partial\ell_\mathbb{E}}{\partial\bm\mu_i}=\mathrm{sgn}\big((\bm\mu_i-\mathbf{\hat{y}}_i)(1+\mathbf{z}_i)\big)\big(1+\mathbf{z}_i\big).
\end{equation*}
Leaving out the non-differentiable $\mathbf{z}_i=-1$, it is easy to prove that
\begin{equation*}
\frac{\partial\ell_\mathbb{E}}{\partial\bm\mu_i}=
\begin{cases}
1+\mathbf{z}_i, & \forall\mathbf{z}_i>-1\\
-1-\mathbf{z}_i, & \text{otherwise}
\end{cases}.
\end{equation*}
In both cases, we always have $\frac{\partial\ell_\mathbb{E}}{\partial\bm\mu_i}>0$. Similarly, for $\bm\mu_i<\mathbf{\hat{y}}_i$, we obtain $\frac{\partial\ell_\mathbb{E}}{\partial\bm\mu_i}<0$ given any $\mathbf{z}_i\sim\mathcal{N}(0,1)$. Recall that $\ell_1$ norm is not differentiable with respect to the origin. Elsewhere, the partial derivative $\mathrm{sgn}(\bm\mu_i-\mathbf{\hat{y}}_i)$ is consistent with the gradient direction of $\frac{\partial\ell_\mathbb{E}}{\partial\bm\mu_i}$ above.

For the estimation of $\Big|\mathbb{E}_\mathbf{z}\big[\frac{\partial \ell_\mathbb{E}}{\partial\bm\mu_i}\big]\Big|$, we shall begin with $\bm\mu_i>\mathbf{\hat{y}}_i$ to simplify the derivation:
\begin{align*}
\mathbb{E}_\mathbf{z}\Big[\frac{\partial\ell_\mathbb{E}}{\partial\bm\mu_i}\Big]&=\int_{-1}^{+\infty}(1+\mathbf{z}_i)\phi(\mathbf{z}_i)\mathrm{d}\mathbf{z}_i-\int^{-1}_{-\infty}(1+\mathbf{z}_i)\phi(\mathbf{z}_i)\mathrm{d}\mathbf{z}_i\\
&=\Phi(+1)-\phi(\mathbf{z}_i)\big\vert_{-1}^{+\infty}-\Phi(-1)+\phi(\mathbf{z}_i)\big\vert^{-1}_{-\infty}\\
&=\Phi(+1)-\Phi(-1)+2\phi(-1)\\
&\approx1.16663
\end{align*}
where $\Phi(\cdot)$ and $\phi(\cdot)$ are the cumulative distribution function (CDF) and probability density function (PDF) of the standard normal distribution respectively. Similarly, we have $\mathbb{E}_\mathbf{z}\big[\frac{\partial \ell_\mathbb{E}}{\partial\bm\mu_i}\big]\approx-1.16663$ for any $\bm\mu_i<\mathbf{\hat{y}}_i$.
\end{proof}

For any $\mathbf{z}\sim\mathcal{N}(0,1)$, we have
\begin{align}
\mathrm{sgn}(\frac{\partial \ell_{\mathbb{E}}}{\partial\bm\mu_i})&=\mathrm{sgn}(\frac{\partial \Vert\mathbf{\hat{y}}-\bm\mu\Vert_1}{\partial\bm\mu_i})\\
\Big|\mathbb{E}_\mathbf{z}\big[\frac{\partial \ell_{\mathbb{E}}}{\partial\bm\mu_i}\big]\Big|&\approx1.
\end{align}
Note that the gradient direction is always making $\bm\mu$ approach $\mathbf{\hat{y}}$ and the step-size converges to nearly one. Therefore, we can safely average the gradient of $\nabla_\mathbf{W}\ell_\mathbb{E}$ over arbitrarily many samplings of $\mathbf{z}$ and training images $\mathbf{x}$, without the risk of gradient ascent or the problem of vanishing/exploding gradient. Due to the ill-posed nature of super-resolution, $|\mathbf{\hat{y}}-\bm\mu|$ can never be a zero vector then the randomness of (\ref{main_loss}) will be preserved.

\begin{figure}[t]
\centering
\includegraphics[width=0.45\textwidth]{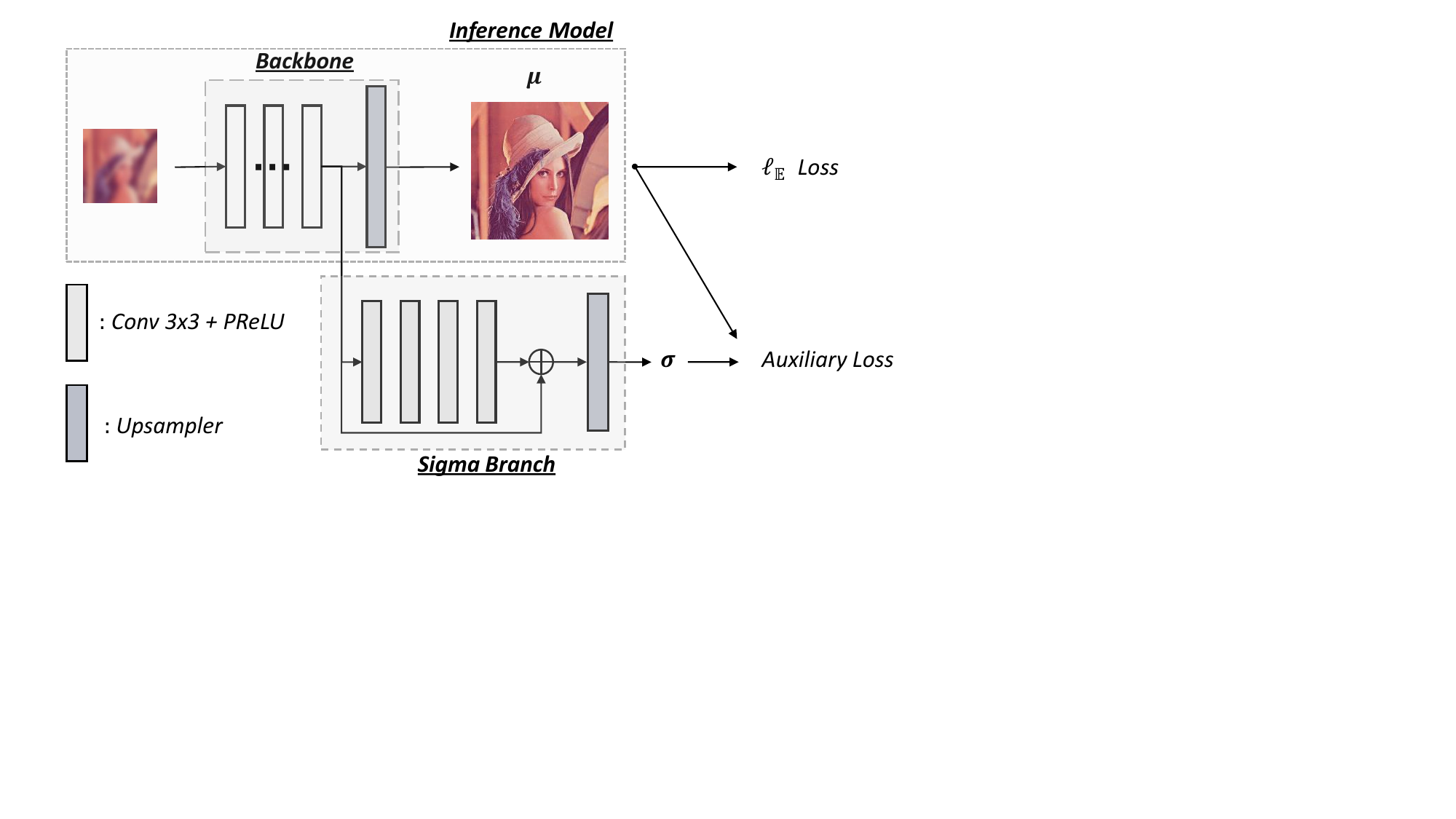}
\caption{An illustration of the proposed learning scheme and the sigma branch. The newly added branch includes two operations: four Conv+PReLU blocks serve as a mapping function and the upsampler (Pixelshuffle coupled with $3\times3$ convolution) adjusts the output size. ``$\ell_\mathbb{E}$'' and ``Auxiliary Loss'' correspond to Equation (\ref{main_loss}) and Equation (\ref{aux_loss}) respectively. We use $\bm\mu$ as the super-resolved image at inference time.}
\label{framework}
\end{figure}

\begin{figure*}[h]
\centering
\begin{subfigure}{.4\textwidth}
\centering
\includegraphics[width=0.99\textwidth]{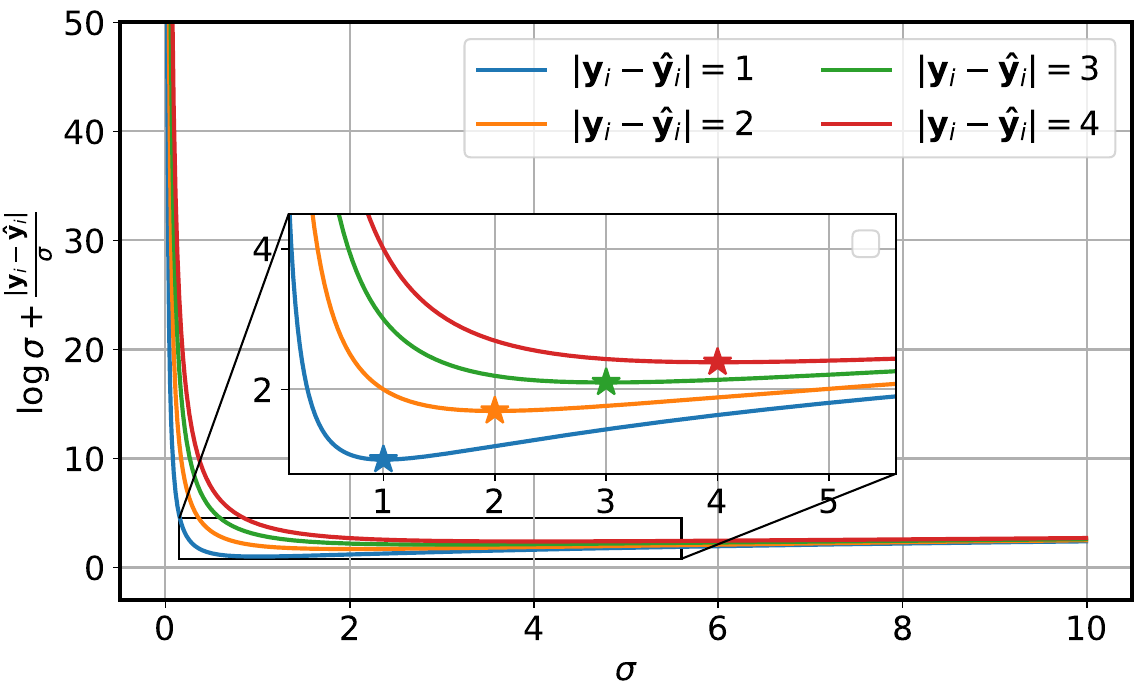}
\caption{Optimization landscape of $\sigma$.}
\end{subfigure}
\begin{subfigure}{.4\textwidth}
\centering
\includegraphics[width=0.99\textwidth]{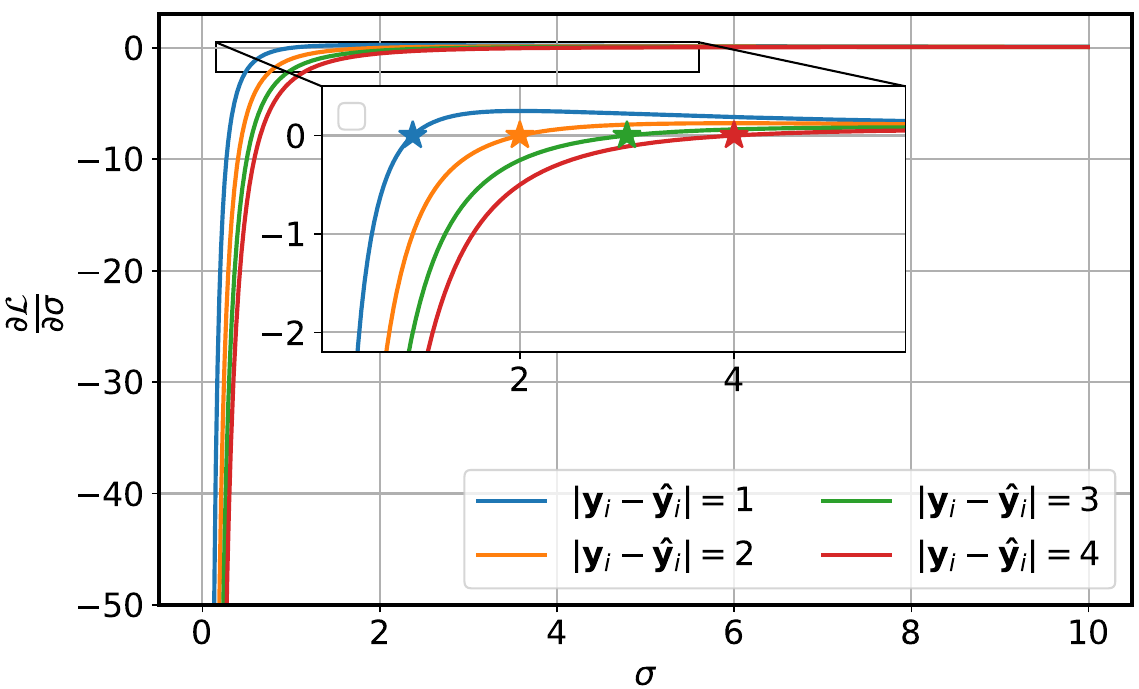}
\caption{Derivative to $\sigma$.}
\label{fig:derivative2sigma}
\end{subfigure}
\caption{Objective function and its derivative to $\sigma$ under the naive setting. The star markers indicate the local optimums.}
\label{fig:object_sigma}
\end{figure*}

It is shown that $|\mathbf{\hat{y}}-\bm\mu|$ plays a core role in learning $\bm\mu$ and serves as $\bm\sigma$ in $\ell_\mathbb{E}$ loss, which inspires us to re-fetch the $\bm\sigma$ branch to perform
the following multi-task learning except $\ell_\mathbb{E}$ loss:
\begin{align}
\min_{\bm\sigma}\ \beta\cdot\big\Vert|\mathbf{\hat{y}}-\bm\mu|-\bm\sigma\big\Vert_1
\label{aux_loss}
\end{align}
where $\beta$ is a penalty factor and $\bm\mu$ is detached from backward propagation. This equation also, as a by-product, allows us to make a rough estimate of the model uncertainty in SISR via the prediction of $\bm\sigma$, \textit{e.g.,} small $\bm\sigma_i$ means we can be certain of the super-resolved image produced by neural networks, otherwise should be less confident (details in Section \ref{uncertainty}). Finally, Figure \ref{framework} further illustrates the overall structure of the proposed scheme.

\subsection{Revisiting Naive Setting}\label{compare_with_naive}
It is intuitive to directly model the joint probability density function $P(\mathbf{y}_i,\mathbf{\hat{y}}_i|\mathbf{x},\mathbf{W})$ in the form of the PDF of Laplician distribution $\frac{1}{2\bm\sigma_i}\exp(-\frac{|\mathbf{y}_i-\mathbf{\hat{y}}_i|}{\bm\sigma_i})$ without considering $P(\mathbf{\hat{y}}_i|\mathbf{y}_i)$. In this subsection, we argue that this naive setting where $\mathbf{y}_i$ and $\bm\sigma_i$ combine into a unified objective function is not suitable for either image restoration or uncertainty estimation.  

Firstly, we are interested in estimating $P(\mathbf{y}_i,\mathbf{\hat{y}}_i|\mathbf{x},\mathbf{W})$ using maximum likelihood estimation
\begin{align}
    \max~\log P(\mathbf{y}_i,\mathbf{\hat{y}}_i|\mathbf{x},\mathbf{W})&=\max~-\log\bm{\sigma}_i-\frac{|\mathbf{y}_i-\mathbf{\hat{y}}_i|}{\bm\sigma_i}\\
    &\Leftrightarrow\min~\log\bm{\sigma}_i+\frac{|\mathbf{y}_i-\mathbf{\hat{y}}_i|}{\bm\sigma_i}
    \label{eq:mle}
\end{align}
where $\mathbf{y}$ is the predicted SR image. It is easy to prove that Equation (\ref{eq:mle}), as a twice differentiable function of $\bm\sigma_i$, is convex on an interval $\bm\sigma_i\in(0,2|\mathbf{y}_i-\mathbf{\hat{y}}_i|]$. The local optimum is surprisingly 
\begin{equation}
    \bm\sigma^*_i=|\mathbf{y}_i-\mathbf{\hat{y}}_i|
\end{equation}
which is consistent with the learning target of Equation (\ref{aux_loss}). However, as illustrated in Figure \ref{fig:object_sigma}, the optimization process can be quite challenging when $|\mathbf{y}_i-\mathbf{\hat{y}}_i|$ is relatively large due to an ill-behaved (more precisely, flat) loss landscape of $\bm\sigma_i$. Note that, the averaged $\ell_1$ training loss on DIV2K dataset is roughly 3.3 in $2\times$ SR (more than 5 for $3\times$ SR), which implies we could not reach $\bm\sigma^*_i$ since the derivative approaches zero along with $\bm\sigma_i\rightarrow\bm\sigma^*_i$, as shown in Figure \ref{fig:derivative2sigma}. In contrast, Equation (\ref{aux_loss}) enjoys a constant derivative almost everywhere with the same $\bm\sigma^*_i$.

Besides, the behavior of Equation (\ref{eq:mle}) can be counter-intuitive when $\bm\sigma_i\rightarrow\bm\sigma^*_i$, which leads to a weighted MAE loss for the prediction of $\mathbf{y}_i$. The gradient then becomes
\begin{equation}
    \frac{\partial\mathcal{L}}{\partial\mathbf{y}_i}=\frac{1}{\bm\sigma_i}\text{sgn}(\mathbf{y}_i-\mathbf{\hat{y}}_i)
\end{equation}
Assuming that $\bm\sigma_i\approx\bm\sigma^*_i$, we may rewrite the above equation into
\begin{equation}
    \frac{\partial\mathcal{L}}{\partial\mathbf{y}_i}\approx\frac{1}{|\mathbf{y}_i-\mathbf{\hat{y}}_i|}\text{sgn}(\mathbf{y}_i-\mathbf{\hat{y}}_i).
\end{equation}
which indicates that \textit{pixels with larger regression error should contribute less to the gradient of $\mathbf{y}$}. To be specific, since the magnitude of $\text{sgn}(\mathbf{y}_i-\mathbf{\hat{y}}_i)$ remains $+1$ for all pixels, the gradient will be dominated by $\frac{1}{|\mathbf{y}_i-\mathbf{\hat{y}}_i|}$. It is common sense that learning based methods should focus on edges with large $|\mathbf{y}_i-\mathbf{\hat{y}}_i|$ instead of signals in low-frequency with small $|\mathbf{y}_i-\mathbf{\hat{y}}_i|$. As for Equation (\ref{eq:mle}), it is just the other way around. Therefore, even if $\bm\sigma_i\rightarrow\bm\sigma^*_i$, one may encounter unexpected results.

\section{Experiments}
\subsection{Dataset}
For Bicubic image SR, following the common setting in \cite{EDSR, OISR, CARN, IGNN, SAN, HAN}, we conduct all experiments on the 1st-800th images from DIV2K \cite{DIV2K} training set then evaluate our models on five benchmarks: Set5 \cite{Set5}, Set14 \cite{Set14}, B100 \cite{B100}, Urban100 \cite{Urban100} and Manga109 \cite{Manga109} with scale factors $2\times$, $3\times$, $4\times$. To make a fair comparison with early works trained on the tiny dataset \cite{A}, we also cite the reproduced DIV2K results \cite{PISR} in Table \ref{tab:cmp}. For real-world SR, we use the same dataset and training settings proposed in RealSR benchmark \cite{RealSR}. We report PSNR and SSIM on the Y channel in YCbCr space and ignore the same amount of pixels as scales from the border. 

Since $\ell_1$ loss is also known-to-work for Additive White Gaussian Noise (AWGN) denoising, we evaluate our methods on another representative task: color (Table \ref{tab:denoising_color_results})/grayscale (Table \ref{tab:denoising_grayscale_results}) image denoising.

\subsection{Training Details}
We use low-resolution RGB image patches with the size of $48\times48$ for training. All HR-LR image pairs are randomly rotated by $90^\circ$, $180^\circ$, $270^\circ$ and flipped horizontally \cite{EDSR}. We set the mini-batch size to 16. Our models are trained by Adam optimizer \cite{adam} with $\beta_1=0.9$, $\beta_2=0.999$, and $\epsilon=10^{-8}$. The initial learning rate is set to $10^{-4}$ and decreases to half at every $2\times10^5$ mini-batch updates. The total training cost is $10^6$ iterations \cite{EDSR}. For RealSR, as described in \cite{RealSR}, the training patch size is $192\times192$ and learning rate is fixed at $10^{-4}$. Other details are the same as classical SR. We use the PyTorch framework \cite{PaszkeGMLBCKLGA19} to implement our methods with an NVIDIA Tesla V100 GPU.

For image denoising, here we use the state-of-the-art DRUNet \cite{DRUNet} as a strong baseline method. Following the same settings in \cite{DRUNet}, we include 400 BSD images \cite{B100}, 4744 images of Waterloo Exploration Database \cite{MaDWWYLZ17}, 900 images from DIV2K \cite{DIV2K}, and 2750 images from Flick2K dataset \cite{EDSR} as the training dataset. The standard AWGN with noise level $\sigma$ is added to the clean image. A specific noise level is randomly chosen from [0,50] during training.  The learning rate
starts from $1e^{-4}$ and finally ends at $1.56e^{-6}$. We set the patch size to $192\times192$ and the batch size is 16 for 4 Nvidia RTX GPUs.

\begin{table}[t]\small
\caption{EDSR-baseline \cite{EDSR} results on Set14 ($2\times$) for different learning targets. We use the best setting in the following experiments.}
\centering
\begin{tabular}{c c c c c c}
\toprule[1.pt]
$\ell_1$ & Eq.(\ref{main_loss}) & Eq.(\ref{aux_loss}) & PSNR (dB) & SSIM & Training cost\\
\hline
\checkmark & & & 33.57 & 0.9175 &\textbf{2.00 ms/iter} \\
& \checkmark & & 33.68 & 0.9181 & 2.06 ms/iter \\
& & \checkmark & 33.57 & 0.9176 & 2.16 ms/iter \\
\checkmark & & \checkmark & 33.64 & 0.9180 & 2.71 ms/iter \\
& \checkmark & \checkmark & \textbf{33.71} & \textbf{0.9185} & 2.77 ms/iter \\
\bottomrule[1.pt]
\end{tabular}
\label{tab:ablation_study1}
\end{table}

\begin{table}[t]\small
\caption{Ablation study on the penalty factor $\beta$. (EDSR-baseline \cite{EDSR} on Set14 $2\times$)}
\centering
\begin{tabular}{c c c c c c}
\toprule[1.pt]
$\beta$ & 0.001 & 0.005 & 0.01 & 0.05 & 0.1 \\
\hline
PSNR (dB) & 33.605 & 33.641 & \textbf{33.707} & 33.663 & 33.647 \\
SSIM & 0.9176 & 0.9181 & \textbf{0.9185} & 0.9184 & 0.9178 \\
\bottomrule[1.pt]
\end{tabular}
\label{tab:ablation_study2}
\end{table}

\begin{figure}[t]
\centering
\includegraphics[width=0.45\textwidth]{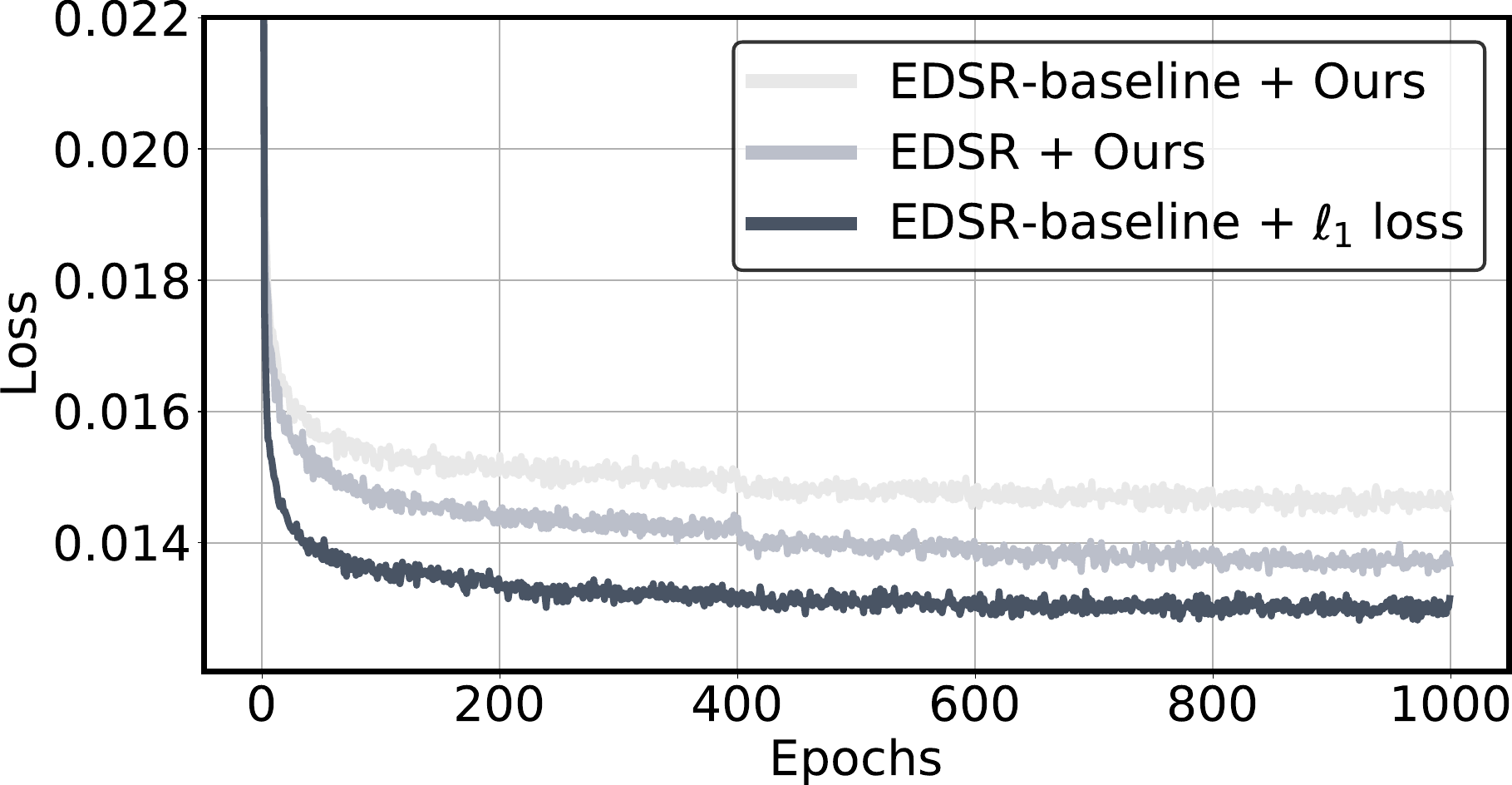}
\caption{The error curves of training $2\times$ EDSR-baseline (1.37M parameters) and $2\times$ EDSR (42M parameters). Our method enjoys nearly the same convergence speed as $\ell_1$ loss. Due to the proposed multi-task learning object, the converged loss value is slightly larger than $\ell_1$ loss.}
\label{loss}
\end{figure}

\subsection{Ablation Studies}
We first present an ablation analysis on each object function in our learning process and report quantitative results in terms of average PSNR and SSIM on Set14 \cite{Set14} with the scale factor of 2. The baseline results in Table \ref{tab:ablation_study1} (\textit{i.e.,} first row) are obtaind using EDSR pre-trained model \footnote[3]{\url{https://cv.snu.ac.kr/research/EDSR/models/edsr_baseline_x2-1bc95232.pt}}. In the case of training neural networks by Equation (\ref{aux_loss}) (\textit{i.e.}, third row), we run the backpropagation though both $\bm\mu$ and $\bm\sigma$ to allow the prediction of super-resolved images via $\bm\mu$ branch. The following rows illustrate that both $\ell_\mathbb{E}$ loss (Equation (\ref{main_loss})) and the multi-task learning (Equation (\ref{aux_loss})) contribute to higher performances than $\ell_1$ baseline. When applying both settings to the original network, we can further improve the PSNR and SSIM results. In the following experiments, we use ``Equation (\ref{main_loss})+Equation (\ref{aux_loss})'' as the learning target.

\begin{figure}[t]\scriptsize
\centering
\begin{subfigure}[b]{.5\textwidth}
\centering
\includegraphics[width=0.92\textwidth]{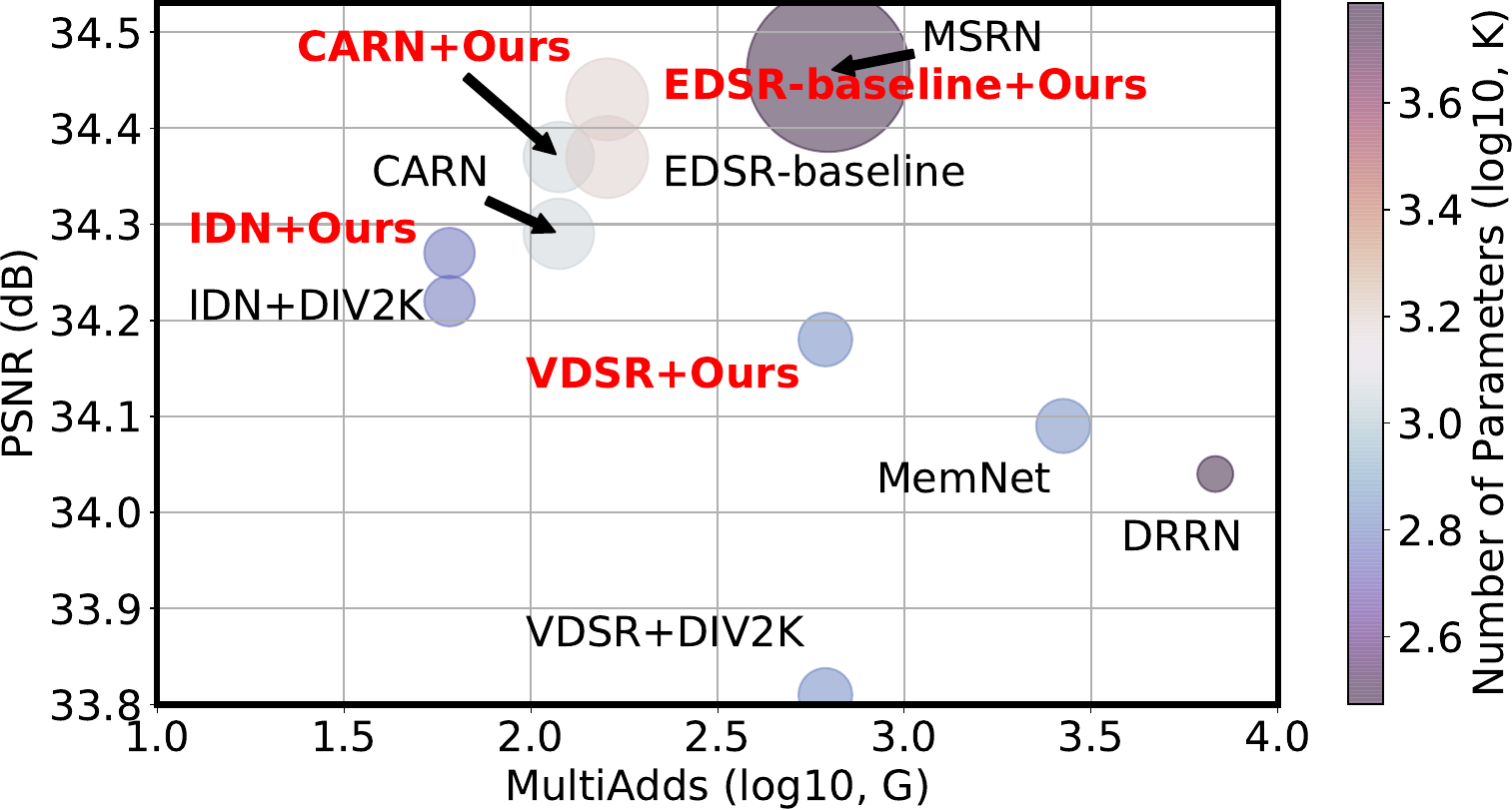}
\caption{Lightweight Models}
\end{subfigure}
\centering
\begin{subfigure}[b]{.5\textwidth}
\centering
\includegraphics[width=0.92\textwidth]{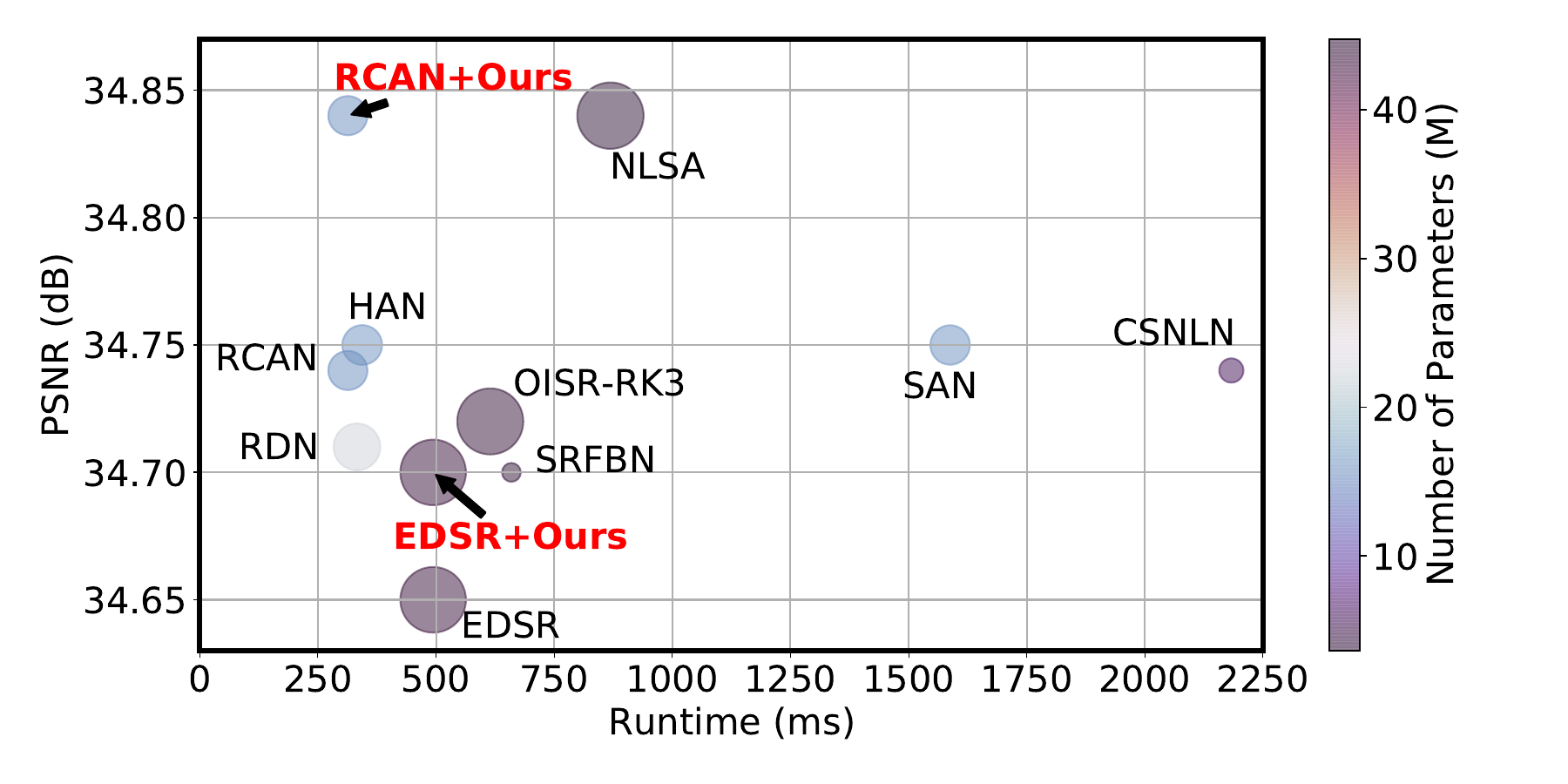}
\caption{Big Models}
\end{subfigure}
\caption{Trade-off between the model size and the average PSNR on Set5 \cite{Set5} (3$\times$). The marker size indicates the number of parameters. The running time is measured on an 720P SR image. (Best viewed in color.)}
\label{tradeoff}
\end{figure}

\begin{table*}[t]\footnotesize
\caption{Quantitative comparison with lightweight models. Our approach improves the performance of early baseline methods to the state-of-the-art. It is shown that networks with simple structures such as EDSR \cite{EDSR} can be competitive with recent elaborate-designed networks. When coupled with efficient models (\textit{e.g.,} CARN \cite{CARN}, IDN \cite{IDN}), the improvement is consistent as shown in Table \ref{tab:cmp}.}
\centering
\begin{tabular}{ l c c c c | c c | c c | c c | c c}
\toprule[1.pt]
\multirow{2}{*}{Scale} & \multirow{2}{*}{Methods} & \multirow{2}{*}{\#Param.} & \multirow{2}{*}{MAC} & \multirow{2}{*}{Runtime} & \multicolumn{2}{c|}{Set5~\cite{Set5}} & \multicolumn{2}{c|}{Set14~\cite{Set14}} & \multicolumn{2}{c|}{B100~\cite{B100}} & \multicolumn{2}{c}{Urban100~\cite{Urban100}} \\
& & & & & PSNR & SSIM & PSNR & SSIM & PSNR & SSIM & PSNR & SSIM \\ \hline
\multirow{7}{*}{2$\times$}
& VDSR \cite{VDSR} & \multirow{2}{*}{668K} & \multirow{2}{*}{614.7G} & \multirow{2}{*}{120.8ms} & 37.53 & 0.9587 & 33.03 & 0.9124 & 31.90 & 0.8960 & 30.76 & 0.9140 \\
& \textbf{VDSR+Ours} & & & & \textbf{37.83} & \textbf{0.9600} & \textbf{33.40} & \textbf{0.9158} & \textbf{32.05} & \textbf{0.8983} & \textbf{31.59} & \textbf{0.9226} \\
\cline{2-13}
& EDSR-baseline \cite{EDSR} & \multirow{2}{*}{1370K} & \multirow{2}{*}{316.2G} & \multirow{2}{*}{56.5ms} & 37.99 & 0.9604 & 33.57 & 0.9175 & 32.16 & 0.8994 & 31.98 & 0.9272 \\
& \textbf{EDSR-baseline+Ours} & & & & \textbf{38.02} & \textbf{0.9606} & \textbf{33.71} & \textbf{0.9185} & \textbf{32.18} & \textbf{0.9001} & \textbf{32.19} & \textbf{0.9292} \\
\cline{2-13}
& \transparent{0.7} CARN \cite{CARN} & \transparent{0.7} 964K & \transparent{0.7} 223.4G & \transparent{0.7} 56.7ms & \transparent{0.7} 37.76 & \transparent{0.7}0.9590 & \transparent{0.7} 33.52 & \transparent{0.7}0.9166 & \transparent{0.7} 32.09 & \transparent{0.7}0.8978 & \transparent{0.7} 31.92 & \transparent{0.7}0.9256 \\
& \transparent{0.7} IMDN \cite{IMDN} & \transparent{0.7} 694K & \transparent{0.7} 159.6G & \transparent{0.7} 43.2ms & \transparent{0.7} 38.00 & \transparent{0.7}0.9605 & \transparent{0.7} 33.63 & \transparent{0.7}0.9177 & \transparent{0.7} 32.19 & \transparent{0.7}0.8996 & \transparent{0.7} 32.17 & \transparent{0.7}0.9283 \\
& \transparent{0.7} OISR \cite{OISR} & \transparent{0.7} 1372K & \transparent{0.7} 316.2G & \transparent{0.7} 68.8ms & \transparent{0.7} 38.02 & \transparent{0.7}0.9605 & \transparent{0.7} 33.62 & \transparent{0.7}0.9178 & \transparent{0.7} 32.20 & \transparent{0.7}0.9000 & \transparent{0.7} 32.21 & \transparent{0.7}0.9290\\
\hline
\multirow{7}{*}{3$\times$}
& VDSR \cite{VDSR} & \multirow{2}{*}{668K} & \multirow{2}{*}{614.7G} & \multirow{2}{*}{117.3ms} & 33.66 & 0.9213 & 29.77 & 0.8314 & 28.82 & 0.7976 & 27.14 & 0.8279 \\
& \textbf{VDSR+Ours} & & & & \textbf{34.18} & \textbf{0.9249} & \textbf{30.18} & \textbf{0.8395} & \textbf{28.99} & \textbf{0.8028} & \textbf{27.80} & \textbf{0.8436} \\
\cline{2-13}
& EDSR-baseline \cite{EDSR} & \multirow{2}{*}{1555K} & \multirow{2}{*}{160.1G} & \multirow{2}{*}{30.4ms} & 34.37 & 0.9270 & 30.28 & 0.8417 & 29.09 & 0.8052 & 28.15 & 0.8527 \\
& \textbf{EDSR-baseline+Ours} & & & & \textbf{34.41} & \textbf{0.9271} & \textbf{30.38} & \textbf{0.8435} & \textbf{29.12} & \textbf{0.8062} & \textbf{28.22} & \textbf{0.8543} \\
\cline{2-13}
& \transparent{0.7} CARN \cite{CARN} & \transparent{0.7} 1149K & \transparent{0.7} 118.9G & \transparent{0.7} 30.7ms & \transparent{0.7} 34.29 & \transparent{0.7}0.9255 & \transparent{0.7} 30.29 & \transparent{0.7}0.8407 & \transparent{0.7} 29.06 & \transparent{0.7}0.8034 & \transparent{0.7} 28.06 & \transparent{0.7}0.8493 \\
& \transparent{0.7} IMDN \cite{IMDN} & \transparent{0.7} 703K & \transparent{0.7} 71.7G & \transparent{0.7} 22.4ms & \transparent{0.7} 34.36 & \transparent{0.7}0.9270 & \transparent{0.7} 30.32 & \transparent{0.7}0.8417 & \transparent{0.7} 29.09 & \transparent{0.7}0.8046 & \transparent{0.7} 28.17 &\transparent{0.7} 0.8519 \\
& \transparent{0.7} OISR \cite{OISR} & \transparent{0.7} 1557K & \transparent{0.7} 160.1G & \transparent{0.7} 36.2ms & \transparent{0.7} 34.39 & \transparent{0.7}0.9272 & \transparent{0.7} 30.35 & \transparent{0.7}0.8426 & \transparent{0.7} 29.11 & \transparent{0.7}0.8058 & \transparent{0.7} 28.24 & \transparent{0.7}0.8544\\
\hline
\multirow{7}{*}{4$\times$}
& VDSR \cite{VDSR} & \multirow{2}{*}{668K} & \multirow{2}{*}{ 614.7G} & \multirow{2}{*}{121.4ms} & 31.35 & 0.8838 & 28.01 & 0.7674 & 27.29 & 0.7251 & 25.18 & 0.7524 \\
& \textbf{VDSR+Ours} & & & & \textbf{31.89} & \textbf{0.8900} & \textbf{28.45} & \textbf{0.7778} & \textbf{27.46} & \textbf{0.7327} & \textbf{25.76} & \textbf{0.7734} \\
\cline{2-13}
& EDSR-baseline \cite{EDSR} & \multirow{2}{*}{1518K} & \multirow{2}{*}{ 114.2G} & \multirow{2}{*}{21.6ms} & 32.09 & \textbf{0.8938} & 28.58 & 0.7813 & 27.57 & 0.7357 & 26.04 & 0.7849 \\
& \textbf{EDSR-baseline+Ours} & & & & \textbf{32.15} & 0.8933 & \textbf{28.65} & \textbf{0.7831} & \textbf{27.59} & \textbf{0.7377} & \textbf{26.15} & \textbf{0.7885} \\
\cline{2-13}
& \transparent{0.7} CARN \cite{CARN} & \transparent{0.7} 1112K & \transparent{0.7} 91.1G & \transparent{0.7} 22.5ms & \transparent{0.7} 32.13 & \transparent{0.7}0.8937 & \transparent{0.7} 28.60 & \transparent{0.7}0.7806 & \transparent{0.7} 27.58 & \transparent{0.7}0.7349 & \transparent{0.7} 26.07 & \transparent{0.7}0.7837 \\
& \transparent{0.7} IMDN \cite{IMDN} & \transparent{0.7} 715K & \transparent{0.7} 41.1G & \transparent{0.7} 11.6ms & \transparent{0.7} 32.21 & \transparent{0.7}0.8948 & \transparent{0.7} 28.58 & \transparent{0.7}0.7811 & \transparent{0.7} 27.56 & \transparent{0.7}0.7353 & \transparent{0.7} 26.04 & \transparent{0.7}0.7838 \\
& \transparent{0.7} OISR \cite{OISR} & \transparent{0.7} 1520K & \transparent{0.7} 114.2G & \transparent{0.7} 24.9ms& \transparent{0.7}32.14 & \transparent{0.7}0.8947 & \transparent{0.7} 28.63 & \transparent{0.7}0.7819 & \transparent{0.7} 27.60 & \transparent{0.7}0.7369 & \transparent{0.7} 26.17 & \transparent{0.7}0.7888 \\
\bottomrule[1.pt]
\end{tabular}
\label{tab:1}
\end{table*}

Table \ref{tab:ablation_study2} indicates that a proper $\beta$ matters in the balance between reconstruction loss $\ell_\mathbb{E}$ and the auxiliary term. 
Fortunately, $0.01$ seems to perform well in most cases. For simplicity, we set the channel number of the sigma branch to 160 and introduce a sigmoid function to normalize its output. In our experiments, we notice that nearly half of the estimated $\bm\sigma$ approach zero, which correspond to the low-frequency/smooth parts of the high-resolution images (as shown in Figure \ref{fig:uncertainty_cmp}). In light of this, we set $|\mathbf{\hat{y}}_i-\bm\mu_i|$ that is less than the average to zero and focus on the hard samples. Due to the limited computing resource, we did not conduct an ablation study on the structure of the sigma branch, instead simply borrowing the idea of Conv+PReLU proposed in \cite{FSRCNN}, shown in Figure \ref{framework}.

\begin{table}[t]\footnotesize
\caption{PSNR comparison with recently proposed loss functions for SISR. PISR \cite{PISR} is a Knowledge Distillation framework where a ``Teacher'' network is required during training. ``DIV2K'' means models reproduced by \cite{PISR} using DIV2K dataset with $\ell_1$ loss. ``EDSR-baseline'' refers to the baseline model in \cite{EDSR}.}
\centering
\resizebox{0.99\linewidth}{!}{
\begin{tabular}{l l c c c}
\toprule[1.pt]
Model & Loss Type & \begin{tabular}[c]{@{}c@{}}2$\times$\\ Set5/B100\end{tabular} & \begin{tabular}[c]{@{}c@{}}3$\times$\\ Set5/B100\end{tabular} & \begin{tabular}[c]{@{}c@{}}4$\times$\\ Set5/B100\end{tabular} \\ \hline
\multirow{6}{*}{VDSR}&
$\ell_2$ & 37.53 / 31.90 & 33.67 / 28.82 & 31.35 / 27.29 \\
& DIV2K~\cite{DIV2K} & 37.64 / 31.96 & 33.80 / 28.83 & 31.37 / 27.25 \\
& Edge~\cite{Edge} & 37.70 / 31.98 & 33.82 / 28.88 & 31.45 / 27.35 \\
& Riemann~\cite{Riemann} & 37.72 / 32.04 & 33.83 / 28.99 & 31.58 / 27.41 \\
& Teacher~\cite{PISR} & 37.77 / 32.00 & 33.85 / 28.86 & 31.51 / 27.29 \\
& \cellcolor{cyan!6}\textbf{Ours} &\cellcolor{cyan!6}\bf{37.83} / \bf{32.05} &\cellcolor{cyan!6}\textbf{34.18} / \bf{28.99} &\cellcolor{cyan!6}\bf{31.89} / \bf{27.46} \\ \hline
\multirow{4}{*}{IDN}&
$\ell_1$ & 37.83 / 32.08 & 34.11 / 28.95 & 31.82 / 27.41 \\
& DIV2K~\cite{DIV2K} & 37.88 / 32.12 & 34.22 / 29.02 & 32.03 / 27.49 \\
& Teacher~\cite{PISR} & \bf{37.93} / \bf{32.14} &34.31 / 29.03 & 32.01 / 27.51 \\
& \cellcolor{cyan!6}\textbf{Ours} &\cellcolor{cyan!6}37.91 / 32.11 & \cellcolor{cyan!6}\textbf{34.33} / \bf{29.04} &\cellcolor{cyan!6}\textbf{32.09} / \bf{27.53} \\\hline
\multirow{3}{*}{CARN}&
$\ell_1$ & 37.76 / 32.09 & 34.29 / 29.06 & 32.13 / 27.58 \\
& Teacher~\cite{PISR} & 37.82 / 32.08 & 34.10 / 28.95 & 31.83 / 27.45 \\
&\cellcolor{cyan!6}\textbf{Ours} &\cellcolor{cyan!6}\bf{37.95} / \bf{32.13} &\cellcolor{cyan!6}\bf{34.37} / \bf{29.08} &\cellcolor{cyan!6}\textbf{32.18 / 27.59} \\\hline
\multirow{3}{*}{\begin{tabular}[l]{@{}l@{}}EDSR-\\baseline\end{tabular}}&
$\ell_1$ & 37.99 / 32.16 & 34.37 / 29.09 & 32.09 / 27.57 \\
& Riemann~\cite{Riemann} & 37.87 / 32.14 & 34.40 / 29.05 & 32.15 / 27.56 \\
&\cellcolor{cyan!6}\textbf{Ours} &\cellcolor{cyan!6}\bf{38.02} / \bf{32.20} &\cellcolor{cyan!6}\textbf{34.43} / \textbf{29.11} &\cellcolor{cyan!6}\textbf{32.15} / \textbf{27.59} \\
\bottomrule[1.pt]
\end{tabular}}
\label{tab:cmp}
\end{table}

\begin{table}[t]
\caption{Average PSNR (dB) for different models on RealSR testing set. We directly cite SRResNet and RCAN results reported in RealSR benchmark \cite{RealSR}. RCAN used in \cite{RealSR} is smaller than the original model. The running time is measured on a $1200\times2200$ SR image.}
\centering
\resizebox{0.97\linewidth}{!}{
\begin{tabular}{l c c c c c }
\toprule[1.pt]
\multirow{2}{*}{Model} & \multirow{2}{*}{\#Param.} & \multirow{2}{*}{Runtime} & \multicolumn{3}{c}{PSNR(dB)} \\
& & & 2$\times$ & 3$\times$ & 4$\times$ \\
\hline
\transparent{0.7}Bicubic & \transparent{0.7}-- & \transparent{0.7}-- & \transparent{0.7}32.61 & \transparent{0.7}29.34 & \transparent{0.7}27.99 \\
\transparent{0.7}DPS \cite{RealSR} & \transparent{0.7}1.26M & \transparent{0.7}122.8ms & \transparent{0.7}32.71 & \transparent{0.7}32.20 & \transparent{0.7}28.69 \\
\transparent{0.7}KPN \cite{RealSR} & \transparent{0.7}1.48M & \transparent{0.7}274.8ms & \transparent{0.7}33.86 & \transparent{0.7}30.39 & \transparent{0.7}28.90 \\
\transparent{0.7}LP-KPN \cite{RealSR} & \transparent{0.7}1.43M & \transparent{0.7}181.3ms & \transparent{0.7}33.90 & \transparent{0.7}30.42 & \transparent{0.7}28.92 \\
\hline
SRResNet \cite{SRGAN}& \multirow{2}{*}{1.31M} & \multirow{2}{*}{143.1ms} &33.69 & 30.18 & 28.67 \\
\textbf{SRResNet+Ours}& & & \textbf{33.87} & \textbf{30.49} & \textbf{28.84} \\
\hline
RCAN \cite{RCAN} & \multirow{2}{*}{8.18M} & \multirow{2}{*}{510.6ms} &33.87 & 30.40 & 28.88 \\
\textbf{RCAN+Ours}& & & \textbf{34.06} & \textbf{30.63} & \textbf{29.00} \\
\bottomrule[1.pt]
\end{tabular}
}
\label{tab:uncertainty}
\end{table}

\subsection{Convergence rate}
Figure \ref{loss} presents a comparison of the convergence rate between EDSR-baseline \cite{EDSR} and standard EDSR model \cite{EDSR}, which indicates that deep SR models are still easy to train compared with lightweight counterparts, using the proposed learning object. Besides, the convergence speed of the proposed scheme is similar to $\ell_1$ loss, though we introduce randomness during training.

\subsection{Quantitative Results}
Figure \ref{tradeoff} shows the performance comparison of our approach coupled with various SR methods to state-of-the-arts in terms of the number of operations and parameters. We notably improve the Pareto frontier with no extra parameter or computing cost at inference time. When compared with recent learning strategies for image restoration, ours performs the best at most cases (Table \ref{tab:cmp}), even surpasses the knowledge distillation method PISR \cite{PISR}. Listed SR methods generally benefit from the proposed scheme except for IDN \cite{IDN} with the scale factor of 2 on B100 \cite{B100}.

\begin{figure*}[t]
\centering
\begin{minipage}[t]{.49\textwidth}
\begin{minipage}[t]{.243\textwidth}
\centering
\includegraphics[width=0.99\textwidth]{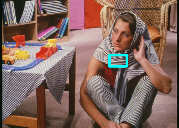}
\includegraphics[width=0.99\textwidth]{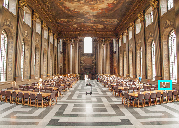}
\includegraphics[width=0.99\textwidth]{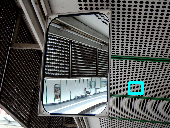}
\subcaption{LR}
\end{minipage}
\begin{minipage}[t]{.243\textwidth}
\centering
\includegraphics[width=0.99\textwidth]{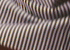}
\includegraphics[width=0.99\textwidth]{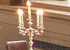}
\includegraphics[width=0.99\textwidth]{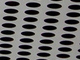}
\subcaption{GT}
\end{minipage}
\begin{minipage}[t]{.243\textwidth}
\centering
\includegraphics[width=0.99\textwidth]{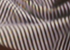}
\includegraphics[width=0.99\textwidth]{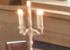}
\includegraphics[width=0.99\textwidth]{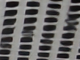}
\subcaption{EDSR \cite{EDSR}}
\label{fig:edsr}
\end{minipage}
\begin{minipage}[t]{.243\textwidth}
\centering
\includegraphics[width=0.99\textwidth]{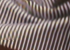}
\includegraphics[width=0.99\textwidth]{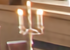}
\includegraphics[width=0.99\textwidth]{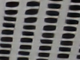}
\subcaption{\cite{EDSR}+Ours}
\end{minipage}
\end{minipage}
\begin{minipage}[t]{.49\textwidth}
\begin{minipage}[t]{.243\textwidth}
\centering
\includegraphics[width=0.99\textwidth]{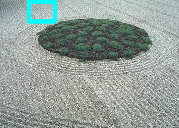}
\includegraphics[width=0.99\textwidth]{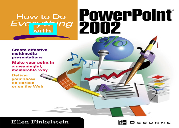}
\includegraphics[width=0.99\textwidth]{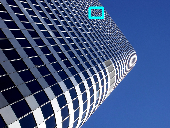}
\subcaption{LR}
\end{minipage}
\begin{minipage}[t]{.243\textwidth}
\centering
\includegraphics[width=0.99\textwidth]{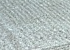}
\includegraphics[width=0.99\textwidth]{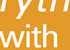}
\includegraphics[width=0.99\textwidth]{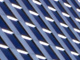}
\subcaption{GT}
\end{minipage}
\begin{minipage}[t]{.243\textwidth}
\centering
\includegraphics[width=0.99\textwidth]{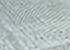}
\includegraphics[width=0.99\textwidth]{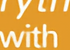}
\includegraphics[width=0.99\textwidth]{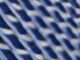}
\subcaption{CARN \cite{CARN}}
\end{minipage}
\begin{minipage}[t]{.243\textwidth}
\centering
\includegraphics[width=0.99\textwidth]{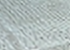}
\includegraphics[width=0.99\textwidth]{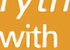}
\includegraphics[width=0.99\textwidth]{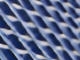}
\subcaption{\cite{CARN}+Ours}
\end{minipage}
\end{minipage}
\caption{Visual comparison for $2\times$, $3\times$ and $4\times$ SR with BI models on benchmark datasets.}
\label{fig:view_cmp}
\end{figure*}

\begin{figure*}[t]
\centering
\begin{minipage}[t]{.115\textwidth}
\centering
\includegraphics[width=0.99\textwidth]{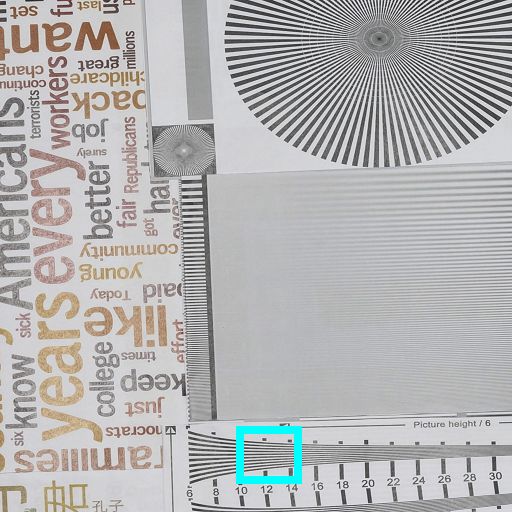}
\includegraphics[width=0.99\textwidth]{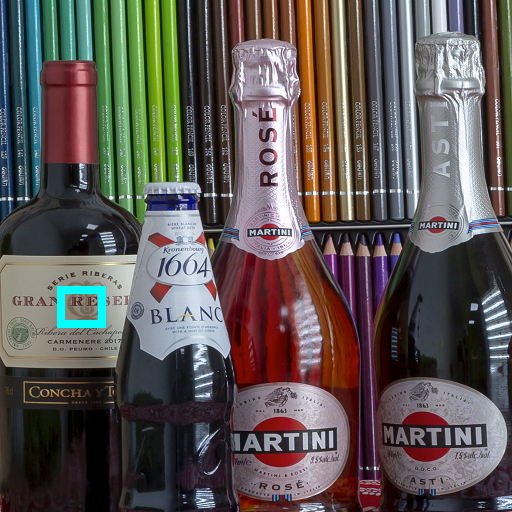}
\subcaption{LR}
\end{minipage}
\begin{minipage}[t]{.115\textwidth}
\centering
\includegraphics[width=0.99\textwidth]{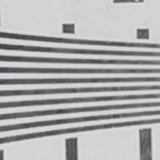}
\includegraphics[width=0.99\textwidth]{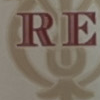}
\subcaption{GT}
\end{minipage}
\begin{minipage}[t]{.115\textwidth}
\centering
\includegraphics[width=0.99\textwidth]{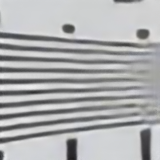}
\includegraphics[width=0.99\textwidth]{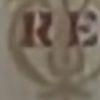}
\subcaption{SRRes\cite{SRGAN}}
\end{minipage}
\begin{minipage}[t]{.115\textwidth}
\centering
\includegraphics[width=0.99\textwidth]{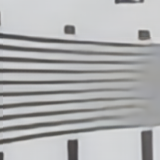}
\includegraphics[width=0.99\textwidth]{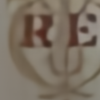}
\subcaption{\cite{SRGAN}+Ours}
\end{minipage}
\begin{minipage}[t]{.115\textwidth}
\centering
\includegraphics[width=0.99\textwidth]{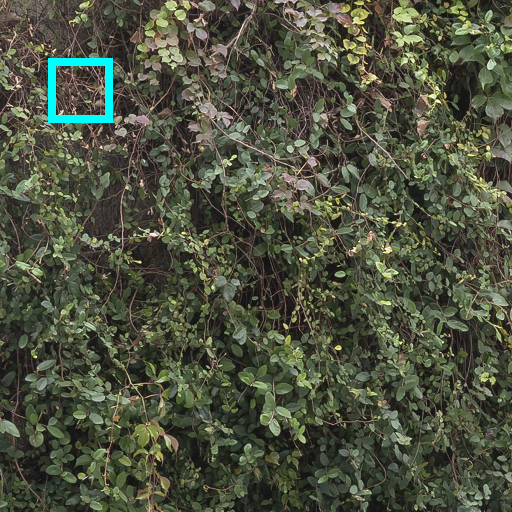}
\includegraphics[width=0.99\textwidth]{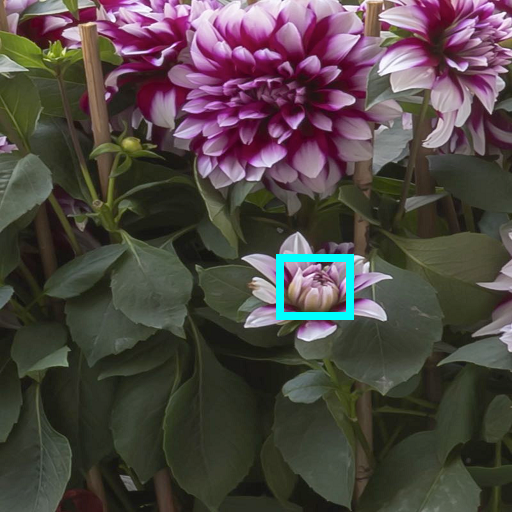}
\subcaption{LR}
\end{minipage}
\begin{minipage}[t]{.115\textwidth}
\centering
\includegraphics[width=0.99\textwidth]{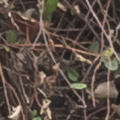}
\includegraphics[width=0.99\textwidth]{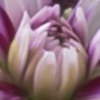}
\subcaption{GT}
\end{minipage}
\begin{minipage}[t]{.115\textwidth}
\centering
\includegraphics[width=0.99\textwidth]{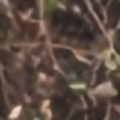}
\includegraphics[width=0.99\textwidth]{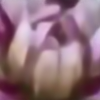}
\subcaption{RCAN\cite{RCAN}}
\end{minipage}
\begin{minipage}[t]{.115\textwidth}
\centering
\includegraphics[width=0.99\textwidth]{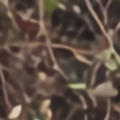}
\includegraphics[width=0.99\textwidth]{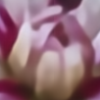}
\subcaption{\cite{RCAN}+Ours}
\end{minipage}
\caption{Qualitative comparison of $3\times$, $4\times$ RealSR results.}
\label{fig:view_cmp_realsr}
\end{figure*}

\begin{table*}[t]\scriptsize
\caption{Quantitative comparison with computing-intensive models. We report the multiply-accumulate operations (MAC) and real runtime cost, required to reconstruct a 1280$\times$720 HR image. RCAN \cite{RCAN} coupled with our methods can be competitive with recent leading works.}
\centering
\resizebox{0.99\linewidth}{!}{
\begin{tabular}{ l c c c | c c | c c | c c | c c | c c}
\toprule[1.pt]
\multirow{2}{*}{Scale} & \multirow{2}{*}{Methods} & \multirow{2}{*}{\#Param.} & \multirow{2}{*}{Runtime} & \multicolumn{2}{c|}{Set5~\cite{Set5}} & \multicolumn{2}{c|}{Set14~\cite{Set14}} & \multicolumn{2}{c|}{B100~\cite{B100}} & \multicolumn{2}{c|}{Urban100~\cite{Urban100}} & \multicolumn{2}{c}{Manga109~\cite{Manga109}} \\
& & & & PSNR & SSIM & PSNR & SSIM & PSNR & SSIM & PSNR & SSIM & PSNR & SSIM \\
\hline
\multirow{7}{*}{2$\times$}
& EDSR \cite{EDSR} & \multirow{2}{*}{40.73M} & \multirow{2}{*}{1003.6ms} & 38.11&0.9602 & 33.92&0.9195 & 32.32&0.9013 & 32.93&0.9351 & 39.10&0.9773 \\
& \textbf{EDSR+Ours} & & & \textbf{38.24}&\textbf{0.9613} & \textbf{34.05}&\textbf{0.9209} & \textbf{32.36}&\textbf{0.9020} & \textbf{32.96}&\textbf{0.9360} & \textbf{39.19}&\textbf{0.9780} \\
\cline{2-14}
& RCAN \cite{RCAN} & \multirow{2}{*}{15.44M} & \multirow{2}{*}{667.8ms} & \textbf{38.27}&0.9614 & 34.12&0.9216 & \textbf{32.41}&\textbf{0.9027} & \textbf{33.34}&\textbf{0.9384} & \textbf{39.44}&\textbf{0.9786} \\
& \textbf{RCAN+Ours} & & & 38.25&\textbf{0.9615} & \textbf{34.16}&\textbf{0.9231} & 32.40&0.9025 & 33.19&0.9381 & 39.41&0.9785 \\
\cline{2-14}
& \transparent{0.7} SAN \cite{SAN} & \transparent{0.7} 15.71M & \transparent{0.7} 4817.6ms & \transparent{0.7} 38.31&\transparent{0.7}0.9620 & \transparent{0.7} 34.07&\transparent{0.7}0.9213 & \transparent{0.7} 32.42&\transparent{0.7}0.9028 & \transparent{0.7} 33.10&\transparent{0.7}0.9370 & \transparent{0.9} 39.32&\transparent{0.7}0.9792 \\
& \transparent{0.7} HAN \cite{HAN} & \transparent{0.7} 15.92M & \transparent{0.7} 743.3ms & \transparent{0.7} 38.27&\transparent{0.7}0.9614 & \transparent{0.7} 34.16&\transparent{0.7}0.9217 & \transparent{0.7} 32.41&\transparent{0.7}0.9027 & \transparent{0.7} 33.35&\transparent{0.7}0.9385 & \transparent{0.9} 39.46&\transparent{0.7}0.9785 \\
& \transparent{0.7} NLSA \cite{NLSA} & \transparent{0.7} 41.80M & \transparent{0.7} 1915.8ms & \transparent{0.7} 38.34&\transparent{0.7}0.9618 & \transparent{0.7} 34.08&\transparent{0.7}0.9231 & \transparent{0.7} 32.43&\transparent{0.7}0.9027 & \transparent{0.7} 33.42&\transparent{0.7}0.9394 & \transparent{0.9} 39.59&\transparent{0.7}0.9789 \\
\hline
\multirow{7}{*}{3$\times$}
& EDSR \cite{EDSR} & \multirow{2}{*}{43.68M} & \multirow{2}{*}{493.8ms} & 34.65&0.9280 & 30.52&0.8462 & 29.25&0.8093 & 28.80&0.8653 & 34.17&0.9476 \\
& \textbf{EDSR+Ours} & & & \textbf{34.70}&\textbf{0.9295} & \textbf{30.58}&\textbf{0.8467} & \textbf{29.28}&\textbf{0.8100} & \textbf{28.87}&\textbf{0.8669} & \textbf{34.18} & \textbf{0.9487} \\
\cline{2-14}
& RCAN \cite{RCAN} & \multirow{2}{*}{15.63M} & \multirow{2}{*}{313.4ms} & 34.74 &0.9299 & 30.65&0.8482 & 29.32& 0.8111 & 29.09 & 0.8702 & 34.44&0.9499 \\
& \textbf{RCAN+Ours} & & & \textbf{34.84}&\textbf{0.9304} & \textbf{30.70}&\textbf{0.8493} & \textbf{29.33}&\textbf{0.8112} & \textbf{29.14}&\textbf{0.8716} & \textbf{34.56}& \textbf{0.9504} \\
\cline{2-14}
& \transparent{0.7} SAN \cite{SAN} & \transparent{0.7} 15.90M & \transparent{0.7} 1,587.9ms & \transparent{0.7} 34.75&\transparent{0.7}0.9300 & \transparent{0.7} 30.59&\transparent{0.7}0.8476 & \transparent{0.7} 29.33&\transparent{0.7}0.8112 & \transparent{0.7} 28.93&\transparent{0.7}0.8671 & \transparent{0.7} 34.30&\transparent{0.7}0.9494 \\
& \transparent{0.7} HAN \cite{HAN} & \transparent{0.7} 16.11M & \transparent{0.7} 343.7ms & \transparent{0.7} 34.75&\transparent{0.7}0.9299 & \transparent{0.7} 30.67&\transparent{0.7}0.8483 & \transparent{0.7} 29.32&\transparent{0.7}0.8110 & \transparent{0.7} 29.10&\transparent{0.7}0.8705 & \transparent{0.7} 34.48&\transparent{0.7}0.9500 \\
& \transparent{0.7} NLSA \cite{NLSA} & \transparent{0.7} 44.75M & \transparent{0.7} 869.0ms & \transparent{0.7} 34.84&\transparent{0.7}0.9306 & \transparent{0.7} 30.70&\transparent{0.7}0.8485 & \transparent{0.7} 29.34&\transparent{0.7}0.8117 & \transparent{0.7} 29.25&\transparent{0.7}0.8726 & \transparent{0.7} 34.57&\transparent{0.7}0.9508 \\
\hline
\multirow{7}{*}{4$\times$}
& EDSR \cite{EDSR} & \multirow{2}{*}{43.10M} & \multirow{2}{*}{322.6ms} & 32.46&0.8968 & 28.80&0.7876 & 27.71&0.7420 & 26.64 &0.8033 & 31.02&0.9148 \\
& \textbf{EDSR+Ours} & & & \textbf{32.48}&\textbf{0.8985} & \textbf{28.86}&\textbf{0.7883} & \textbf{27.74}&\textbf{0.7423} & \textbf{26.68}&\textbf{0.8045} & \textbf{31.12} & \textbf{0.9165} \\
\cline{2-14}
& RCAN \cite{RCAN} & \multirow{2}{*}{15.59M} & \multirow{2}{*}{181.1ms} & 32.63 & 0.9002 & 28.87&0.7889 & 27.77 &0.7436 & 26.82&0.8087 & 31.22&0.9173 \\
& \textbf{RCAN+Ours} & & & \textbf{32.69}& \textbf{0.9005} & \textbf{28.89} & \textbf{0.7892} & \textbf{27.78}& \textbf{0.7437} & \textbf{26.87} & \textbf{0.8094} & \textbf{31.40} & \textbf{0.9185} \\
\cline{2-14}
& \transparent{0.7} SAN \cite{SAN} & \transparent{0.7} 15.86M & \transparent{0.7} 880.0ms & \transparent{0.7} 32.64&\transparent{0.7}0.9003 & \transparent{0.7} 28.92&\transparent{0.7}0.7888 & \transparent{0.7} 27.78&\transparent{0.7}0.7436 & \transparent{0.7} 26.79&\transparent{0.7}0.8068 & \transparent{0.7} 31.18&\transparent{0.7}0.9169 \\
& \transparent{0.7} HAN \cite{HAN} & \transparent{0.7} 16.07M & \transparent{0.7} 196.8ms & \transparent{0.7} 32.64&\transparent{0.7}0.9002 & \transparent{0.7} 28.90&\transparent{0.7}0.7890 & \transparent{0.7} 27.80&\transparent{0.7}0.7442 & \transparent{0.7} 26.85&\transparent{0.7}0.8094 & \transparent{0.7} 31.42&\transparent{0.7}0.9177 \\
& \transparent{0.7} NLSA \cite{NLSA} & \transparent{0.7} 44.15M & \transparent{0.7} 523.6ms & \transparent{0.7} 32.59&\transparent{0.7}0.9000 & \transparent{0.7} 28.87&\transparent{0.7}0.7891 & \transparent{0.7} 27.78&\transparent{0.7}0.7444 & \transparent{0.7} 26.96&\transparent{0.7}0.8109 &\transparent{0.7} 31.27 &\transparent{0.7}0.9184 \\
\bottomrule[1.pt]
\end{tabular}}
\label{tab:2}
\end{table*}

We further detail the performance of our approach in contrast to recent lightweight networks \cite{CARN,OISR,IMDN} in Table \ref{tab:1}. To well describe different methods, we report the number of parameters, operations required to reconstruct a 720P image, and its average runtime speed measured on an NVIDIA Titan RTX GPU. From Table \ref{tab:1}, we observe that VDSR \cite{VDSR} trained by the proposed learning process outperforms baselines by a large margin, consistently for all scaling factors. Former state-of-the-art EDSR-baseline \cite{EDSR} with our approach also shows competitive results with recent well-designed lightweight SR methods. We further conduct experiments on the large BI degradation model EDSR \cite{EDSR}. As shown in Table \ref{tab:2}, the original model has achieved high PSNR/SSIM on each dataset, while ours can further enhance the performance in most cases. For $3\times$ and $4\times$ Manga109 in particular, the PSNR gains of ours over RCAN are 0.112dB and 0.18 dB. 

To further verify the effectiveness of our approach, we conduct extensive experiments on image denoising datasets with a strong baseline method DRUNet \cite{DRUNet}. Since DRUNet has achieved promising results, we believe the improvements can be nontrivial. Note that, our methods outperform baseline with $\sim0.1$dB on Urban100 and consistently bypass the previous leading methods on all datasets. The results are listed in Table \ref{tab:denoising_grayscale_results} and \ref{tab:denoising_color_results}.

\begin{table*}[!t]\scriptsize
\center
\begin{center}
\caption{Quantitative comparison (average PSNR reported) with state-of-the-art methods for grayscale denoising on benchmark datasets. Best and second best results are shown in \R{red} and \B{blue} colors, respectively. For DRUNet \cite{DRUNet}, we use the official well-trained model as a strong baseline \protect\footnotemark. All testing datasets are the same as FFDNet \cite{zhang2018ffdnet}.}
\label{tab:denoising_grayscale_results}
\begin{tabular}
{c|c|c|c|c|c|c|c|c|c|c|c|c|c|c}
\toprule[1.pt]
~Dataset~ & ~$\sigma$~ & \makecell{BM3D\\\cite{dabov2007bm3d}} & \makecell{WNNM\\\cite{gu2014weighted}} & \makecell{DnCNN\\\cite{zhang2017DnCNN}} & 
\makecell{IRCNN\\\cite{zhang2017IRCNN}} &
\makecell{FFDNet\\\cite{zhang2018ffdnet}} &
\makecell{N$^3$Net\\\cite{plotz2018N3Net}} &  \makecell{NLRN\\\cite{liu2018NLRN}} &
\makecell{FOCNet\\\cite{jia2019focnet}} &
\makecell{RNAN\\\cite{zhang2019RNAN}} &  
\makecell{MWCNN\\\cite{liu2018MWCNN}} & 
\makecell{GCDN\\\cite{GCDN}} & 
\makecell{DRUNet\\\cite{zhang2021DPIR}} &
\makecell{\textbf{\cite{zhang2021DPIR}+Ours}}
\\
\hline
\multirow{3}{*}{\makecell{Set12\\\cite{zhang2017DnCNN}}} & 15
& 32.37 
& 32.70
& 32.86
& 32.76 
& 32.75
& -
& 33.16 
& 33.07
& -
& 33.15
& 33.14
& \B{33.25}
& \R{33.29}
\\
& 25
& 29.97
& 30.28
& 30.44
& 30.37
& 30.43
& 30.55
& 30.80
& 30.73
& -
& 30.79
& 30.78
& \B{30.94}
& \R{30.99}
\\
& 50
& 26.72
& 27.05
& 27.18
& 27.12
& 27.32 
& 27.43
& 27.64
& 27.68
& 27.70
& 27.74
& 27.60
& \B{27.90}
& \R{27.95}
\\
\hline
\multirow{3}{*}{\makecell{BSD68\\\cite{BSD68}}} & 15
& 31.08
& 31.37
& 31.73
& 31.63
& 31.63
& -
& 31.88
& 31.83
& -
& 31.86
& 31.83
& \B{31.90}
& \R{31.93}
\\
& 25
& 28.57
& 28.83
& 29.23
& 29.15
& 29.19
& 29.30
& 29.41
& 29.38
& -
& 29.41
& 29.35
& \B{29.46}
& \R{29.49}
\\
& 50
& 25.60
& 25.87
& 26.23
& 26.19
& 26.29
& 26.39
& 26.47
& 26.50
& 26.48
& 26.53
& 26.38
& \B{26.57}
& \R{26.59}
\\
\hline
\multirow{3}{*}{\makecell{\hspace{-0.15cm}Urban100\\\cite{Urban100}}} & 15
& 32.35
& 32.97
& 32.64
& 32.46
& 32.40
& -
& 33.45
& 33.15
& -
& 33.17
& \B{33.47}
& 33.44
& \R{33.50}
\\
& 25
& 29.70
& 30.39
& 29.95
& 29.80
& 29.90
& 30.19
& 30.94
& 30.64
& -
& 30.66
& 30.95
& \B{31.10}
& \R{31.17}
\\
& 50
& 25.95
& 26.83
& 26.26
& 26.22
& 26.50
& 26.82
& 27.49
& 27.40
& 27.65
& 27.42
& 27.41
& \B{27.96}
& \R{28.03}
\\
\bottomrule[1.pt]       
\end{tabular}
\end{center}
\end{table*}

\begin{table*}[t]\scriptsize
\center
\begin{center}
\caption{Quantitative comparison (average PSNR reported) with state-of-the-art methods for color image denoising on benchmark datasets. Best and second best results are shown in \R{red} and \B{blue} colors, respectively. For DRUNet \cite{DRUNet}, we use the official well-trained model as a strong baseline \protect\footnotemark. All testing datasets are the same as FFDNet \cite{zhang2018ffdnet}.}
\label{tab:denoising_color_results}
\begin{tabular}{c|c|c|c|c|c|c|c|c|c|c|c|c|c}
\toprule[1.pt]
~~~~~Dataset~~~~~ & ~$\sigma$~ & \makecell{BM3D\\\cite{dabov2007bm3d}} & \makecell{DnCNN\\\cite{zhang2017DnCNN}} &
\makecell{IRCNN\\\cite{zhang2017IRCNN}} &
\makecell{FFDNet\\\cite{zhang2018ffdnet}} &
\makecell{RIDNet\\\cite{RIDNet}} & 
\makecell{RPCNN\\\cite{xia2020rpcnn}} & 
\makecell{BRDNet\\\cite{tian2020BRDnet}} & 
\makecell{RNAN\\\cite{RNAN}} &
\makecell{RDN\\\cite{RDN}} &
\makecell{IPT\\\cite{chen2021IPT}} &
\makecell{DRUNet\\\cite{DRUNet}} &
\makecell{\textbf{\cite{DRUNet}+Ours}}
\\
\hline
\multirow{3}{*}{\makecell{CBSD68\\\cite{BSD68}}} & 15
& 33.52   
& 33.90
& 33.86
& 33.87
& 34.01
& -
& 34.10
& -
& -
& -
& \B{34.29}
& \R{34.33}
\\
& 25
& 30.71  
& 31.24 
& 31.16
& 31.21
& 31.37
& 31.24
& 31.43
& -
& -
& -
& \B{31.67}
& \R{31.72}
\\
& 50
& 27.38   
& 27.95
& 27.86
& 27.96
& 28.14
& 28.06
& 28.16
& 28.27
& 28.31
& 28.39
& \B{28.48}
& \R{28.53}
\\
\hline
\multirow{3}{*}{\makecell{Kodak24\\\cite{franzen1999kodak}}} & 15
& 34.28  
& 34.60 
& 34.69
& 34.63
& -
& -
& 34.88
& -
& -
& -
& \B{35.18}
& \R{35.23}
\\
& 25
& 32.15   
& 32.14
& 32.18
& 32.13
& -
& 32.34
& 32.41
& -
& -
& -
& \B{32.76}
& \R{32.81}
\\
& 50
& 28.46   
& 28.95
& 28.93
& 28.98
& -
& 29.25
& 29.22
& 29.58
& 29.66
& 29.64
& \B{29.71}
& \R{29.75}
\\
\hline
\multirow{3}{*}{\makecell{McMaster\\\cite{zhang2011McMaster}}} & 15
& 34.06   
& 33.45
& 34.58
& 34.66
& -
& -
& 35.08
& -
& -
& -
& \B{35.39}
& \R{35.41}
\\
& 25
& 31.66  
& 31.52
& 32.18
& 32.35
& -
& 32.33
& 32.75
& -
& -
& -
& \B{33.13}
& \R{33.16}
\\
& 50
& 28.51  
& 28.62 
& 28.91
& 29.18
& -
& 29.33
& 29.52
& 29.72
& -
& 29.98
& \B{30.07}
& \R{30.10}
\\
\hline
\multirow{3}{*}{\makecell{Urban100\\\cite{Urban100}}} & 15
& 33.93
& 32.98
& 33.78
& 33.83
& -
& -
& 34.42
& -
& -
& -
& \B{34.81}
& \R{34.86}
\\
& 25
& 31.36
& 30.81
& 31.20
& 31.40
& -
& 31.81
& 31.99
& -
& -
& -
& \B{32.60}
& \R{32.67}
\\
& 50
& 27.93
& 27.59
& 27.70
& 28.05
& -
& 28.62
& 28.56
& 29.08
& 29.38
& \R{29.71}
& {29.60}
& \B{29.69}
\\
\bottomrule[1.pt]  
\end{tabular}
\end{center}
\end{table*}

\begin{table}[t]
\caption{The quality of model uncertainty measured on test datasets. PLL (higher is better $\uparrow$) is the widely accepted uncertainty metric. Since uncertainty should reflect underlying mistakes in super-resolved images, PSNR ($\uparrow$) between the predicted uncertainy $\bm\sigma$ and the actual fitting error $|\mathbf{\hat{y}}-\bm\mu|$ is also reported (\textit{i.e.}, PSNR between Fig. \ref{subfigure1} and Fig. \ref{subfigure2}).}
\centering
\resizebox{\linewidth}{!}{
\begin{tabular}{l c c c c c}
\toprule[1.pt]
Model & Scale & \begin{tabular}[c]{@{}c@{}}S5~\cite{Set5}\\ PLL/PSNR\end{tabular} & \begin{tabular}[c]{@{}c@{}}S14~\cite{Set14}\\ PLL/PSNR\end{tabular} & \begin{tabular}[c]{@{}c@{}}B100~\cite{B100}\\ PLL/PSNR\end{tabular} & \begin{tabular}[c]{@{}c@{}}U100~\cite{Urban100}\\ PLL/PSNR\end{tabular}\\
\hline
\multirow{3}{*}{\transparent{0.9}EDSR} & \transparent{0.9}{2$\times$} & \transparent{0.9}2.60 / 39.08 & \transparent{0.9}1.50 / 35.73 & \transparent{0.9}1.08 / 33.70 & \transparent{0.9} 1.32 / 34.09 \\
& \transparent{0.9}3$\times$ & \transparent{0.9}2.28 / 35.82 & \transparent{0.9} 1.21 / 32.28 & \transparent{0.9}0.86 / 30.65 & \transparent{0.9}0.74 / 29.98 \\
& \transparent{0.9}4$\times$ & \transparent{0.9}2.06 / 33.67 & \transparent{0.9} 1.12 / 30.44 & \transparent{0.9}0.87 / 29.15 & \transparent{0.9}0.52 / 27.84 \\
\hline
\cellcolor{cyan!6}& \cellcolor{cyan!6} 2$\times$ & \cellcolor{cyan!6}2.86 / 41.92 & \cellcolor{cyan!6} 2.04 / 37.43 & \cellcolor{cyan!6} 2.25 / 35.86 & \cellcolor{cyan!6} 2.03 / 36.17 \\
\cellcolor{cyan!6}& \cellcolor{cyan!6} 3$\times$ & \cellcolor{cyan!6}2.43 / 38.27 & \cellcolor{cyan!6} 1.55 / 33.93 & \cellcolor{cyan!6} 1.83 / 32.67 & \cellcolor{cyan!6} 1.31 / 31.81 \\
\cellcolor{cyan!6}\multirow{-3}{*}{\begin{tabular}[c]{@{}c@{}}EDSR\\+Ours\end{tabular}\cellcolor{cyan!6}} & \cellcolor{cyan!6} 4$\times$ & \cellcolor{cyan!6} 1.98 / 35.87 & \cellcolor{cyan!6} 1.03 / 31.93 & \cellcolor{cyan!6} 1.49 / 30.95 & \cellcolor{cyan!6} 0.78 / 29.33 \\
\hline
\multirow{3}{*}{\transparent{0.9}CARN} & \transparent{0.9}{2$\times$} & \transparent{0.9}2.58 / 38.56 & \transparent{0.9}1.54 / 35.28 & \transparent{0.9}1.19 / 33.47 & \transparent{0.9}1.14 / 32.96 \\
& \transparent{0.9}3$\times$ & \transparent{0.9}2.25 / 35.24 & \transparent{0.9}1.28 / 31.94 & \transparent{0.9}0.99 / 30.45 & \transparent{0.9}0.70 / 29.28 \\
& \transparent{0.9}4$\times$ & \transparent{0.9}2.02 / 33.23 & \transparent{0.9}1.13 / 30.17 & \transparent{0.9}0.91 / 29.01 & \transparent{0.9}0.44 / 27.35 \\
\hline
\cellcolor{cyan!6} & \cellcolor{cyan!6} 2$\times$ & \cellcolor{cyan!6} 2.80 / 41.67 & \cellcolor{cyan!6} 1.98 / 36.99 & \cellcolor{cyan!6} 2.29 / 35.70 & \cellcolor{cyan!6} 2.07 / 35.35 \\
\cellcolor{cyan!6} & \cellcolor{cyan!6} 3$\times$ & \cellcolor{cyan!6} 2.28 / 37.92 & \cellcolor{cyan!6} 1.40 / 33.62 & \cellcolor{cyan!6} 1.80 / 32.46 & \cellcolor{cyan!6} 1.34 / 31.19 \\
\cellcolor{cyan!6} \multirow{-3}{*}{\begin{tabular}[c]{@{}c@{}}CARN\\+Ours\end{tabular}} & \cellcolor{cyan!6} 4$\times$ & \cellcolor{cyan!6} 2.07 / 35.80 & \cellcolor{cyan!6} 1.17 / 31.89 & \cellcolor{cyan!6} 1.60 / 30.97 & \cellcolor{cyan!6} 0.99 / 29.11 \\
\bottomrule[1.pt]
\end{tabular}
}
\label{tab:uncertainty}
\end{table}

\subsection{Qualitative Results}
To further illustrate the analysis above, we show visual comparisons in Figure \ref{fig:view_cmp} and \ref{fig:view_cmp_realsr}. Bicubic downsampling results in the loss of texture details and structure information, which induces super-resolved images with blurring and artifacts. Even worse, EDSR generates perceptually convincing results such as ``barbara'' in Set14 (first row in Figure \ref{fig:edsr}). However, the SR image contains several lines with wrong directions. Instead, our models can recover them being more faithful to the ground truth. CARN \cite{CARN} with ours also obtains much better results by recovering more informative details.

Despite the promising results, networks can make mistakes when exposed to unseen data outside the trained distribution. In this case, we expect information about model uncertainty in addition to the SR images.

\footnotetext{\url{https://drive.google.com/file/d/1oSsLjPPn6lqtzraFZLZGmwP_5KbPfTES/view?usp=sharing}}
\footnotetext{\url{https://drive.google.com/file/d/1KDn0ok5Q6dJtAAIBBkiFbHl1ms9kVezz/view?usp=sharing}}

\subsection{Model Uncertainty}
\label{uncertainty}
\begin{figure}[t]
\centering
\begin{minipage}[t]{.09\textwidth}
\centering
\includegraphics[width=0.99\textwidth]{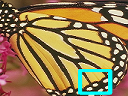}
\includegraphics[width=0.99\textwidth]{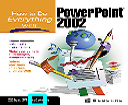}
\includegraphics[width=0.99\textwidth]{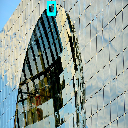}
\includegraphics[width=0.99\textwidth]{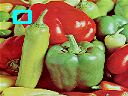}
\includegraphics[width=0.99\textwidth]{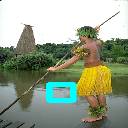}
\includegraphics[width=0.99\textwidth]{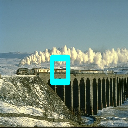}
\includegraphics[width=0.99\textwidth]{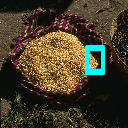}
\includegraphics[width=0.99\textwidth]{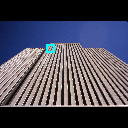}
\subcaption{LR}
\end{minipage}
\begin{minipage}[t]{.09\textwidth}
\centering
\includegraphics[width=0.99\textwidth]{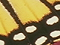}
\includegraphics[width=0.99\textwidth]{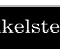}
\includegraphics[width=0.99\textwidth]{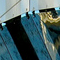}
\includegraphics[width=0.99\textwidth]{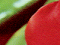}
\includegraphics[width=0.99\textwidth]{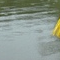}
\includegraphics[width=0.99\textwidth]{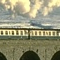}
\includegraphics[width=0.99\textwidth]{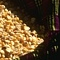}
\includegraphics[width=0.99\textwidth]{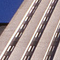}
\subcaption{HR: $\mathbf{\hat{y}}$}
\end{minipage}
\begin{minipage}[t]{.09\textwidth}
\centering
\includegraphics[width=0.99\textwidth]{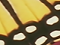}
\includegraphics[width=0.99\textwidth]{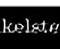}
\includegraphics[width=0.99\textwidth]{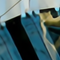}
\includegraphics[width=0.99\textwidth]{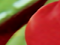}
\includegraphics[width=0.99\textwidth]{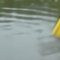}
\includegraphics[width=0.99\textwidth]{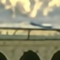}
\includegraphics[width=0.99\textwidth]{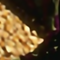}
\includegraphics[width=0.99\textwidth]{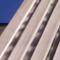}
\subcaption{SR: $\bm\mu$}
\end{minipage}
\begin{minipage}[t]{.09\textwidth}
\centering
\includegraphics[width=0.99\textwidth]{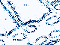}
\includegraphics[width=0.99\textwidth]{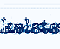}
\includegraphics[width=0.99\textwidth]{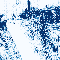}
\includegraphics[width=0.99\textwidth]{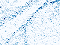}
\includegraphics[width=0.99\textwidth]{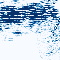}
\includegraphics[width=0.99\textwidth]{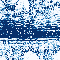}
\includegraphics[width=0.99\textwidth]{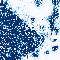}
\includegraphics[width=0.99\textwidth]{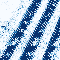}
\subcaption{$|\mathbf{\hat{y}}-\bm\mu|$}
\label{subfigure2}
\end{minipage}
\begin{minipage}[t]{.09\textwidth}
\centering
\includegraphics[width=0.99\textwidth]{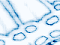}
\includegraphics[width=0.99\textwidth]{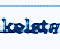}
\includegraphics[width=0.99\textwidth]{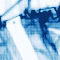}
\includegraphics[width=0.99\textwidth]{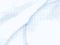}
\includegraphics[width=0.99\textwidth]{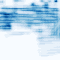}
\includegraphics[width=0.99\textwidth]{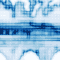}
\includegraphics[width=0.99\textwidth]{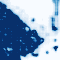}
\includegraphics[width=0.99\textwidth]{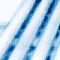}
\subcaption{$\bm\sigma$}
\label{subfigure1}
\end{minipage}
\caption{The last column (\ref{subfigure1}) visualizes the estimated uncertainty $\bm\sigma$ in $3\times$ SR. Dark color indicates large variance, \textit{i.e.,} high uncertainty. Considering the visual differences between HR and SR images and rael residual error (\ref{subfigure2}), $\bm\sigma$ depicts a promising uncertainty estimation that approximates the error with high fidelity.}
\label{fig:uncertainty_cmp}
\end{figure}

By introducing a posterior distribution $P(\mathbf{y}|\mathbf{x};\mathbf{W})$, we can easily use the predicted $\bm\sigma$ as the estimated model uncertainty (Figure \ref{subfigure1}). Following previous Bayesian models \cite{Hernandez, GalG16, TeyeAS18, BuiHHLT16}, we evaluate uncertainty quality based on the standard metric Predictive Log Likelihood (PLL):
\begin{equation}
\mathrm{PLL}(f(\mathbf{x}), (\mathbf{\hat{y}},\mathbf{x}))=\log P(\mathbf{\hat{y}}|f(\mathbf{x}))
\end{equation}
where $P(\mathbf{\hat{y}}|f(\mathbf{x}))$ is the probability of target $\mathbf{\hat{y}}$ generated by a probalistic model $f(\mathbf{x})$. In this paper, $f(\mathbf{x})$ corresponds to the multivariate Gaussian $\mathcal{N}(\bm\mu,\bm\sigma^2)$ with predicted mean $\bm\mu$ and standard deviation $\bm\sigma$. $P(\mathbf{\hat{y}}|f(\mathbf{x}))$ is the Gaussian probability density function (PDF) evaluated at $\mathbf{\hat{y}}$.

We report the average PLL of our approach coupled with EDSR \cite{EDSR} and CARN \cite{CARN} in Table \ref{tab:uncertainty}. The original EDSR/CARN networks fail to consider the model uncertainty, hence we use their average fitting error $\mathbb{E}_\mathcal{D}[|\mathbf{\hat{y}}-\bm\mu|]$ calculated on the training dataset $\mathcal{D}$ to create a constant uncertainty estimation, which corresponds to baseline results in Table \ref{tab:uncertainty}. Any improvement over the baseline reflects a sensible estimate of uncertainty. It is shown that PLL for high upscaling factors are generally low, which indicates that reconstructed $4\times$ image itself can be unreliable, \textit{i.e.}, networks fail to recognize the hidden mistakes.

To improve the interpretability of the model uncertainty, we visualize the predicted $\bm\sigma$ in Figure \ref{subfigure1}. It is shown that the estimated uncertainty strongly correlates with the actual residual error $|\mathbf{\hat{y}}-\bm\mu|$. We also report the PSNR results between them to further evaluate the uncertainty quality fairly since PLL has been criticized for being vulnerable to outliers \cite{selten1998axiomatic}.

\section{Implementation}
In this section, we provide Python-style pseudo-codes to facilitate the implementation and further understanding of our approach by automatic differentiation, which has been well supported in mainstream deep-learning frameworks. 

As discussed in Section \ref{sect:degrad}, the main limitation of $\ell_1$ loss is the degeneracy problem. By recalling Equation (\ref{main_loss})
\begin{align*}
\min_{\bm\mu}\ \mathbb{E}_\mathbf{z}\Big[\big\Vert\mathbf{\hat{y}}-(\bm\mu+\underbrace{|\mathbf{\hat{y}}-\bm\mu|*\mathbf{z}}_{\text{training variance}})\big\Vert_1\Big],
\end{align*}
as $|\mathbf{\hat{y}}-\bm\mu|\rightarrow0$, the variance term approaches zero then the proposed Equation (\ref{main_loss}) also gradually degrades into $\ell_1$ loss
\begin{align*}
\min_{\bm\mu}\ \mathbb{E}_\mathbf{z}\Big[\big\Vert\mathbf{\hat{y}}-(\bm\mu+0^+*\mathbf{z})\big\Vert_1\Big].
\end{align*}
Note that large models lead to much better results than lightweight models, \textit{i.e.,} the residual error $|\mathbf{\hat{y}}-\bm\mu|$ is much smaller in practice, which indicates that cumbersome models may not benefit as much as lightweight ones from Equation (\ref{main_loss}). 

To resolve this issue, we present a simple heuristic method that only applys Equation (\ref{main_loss}) to ``hard samples'' with large $|\mathbf{\hat{y}}_i-\bm\mu_i|$. This setting shares a similar idea with Focal-Loss \cite{focal_loss} and prevents the vast number of easy samples (\textit{i.e.,} $|\mathbf{\hat{y}}_i-\bm\mu_i|\approx0$) from overwhelming the sigma branch during training.
\begin{algorithm}[!]
\caption{Pseudo-codes of Equation (\ref{main_loss},\ref{aux_loss}) in a Python style.}
\label{alg:code}
\algcomment{\fontsize{7.5pt}{0em}\selectfont \texttt{*}: element-wise product.
}
\definecolor{codeblue}{rgb}{0.25,0.5,0.5}
\definecolor{codered}{rgb}{0.75,0.25,0.25}
\definecolor{codepurple}{rgb}{0.58,0,0.82}
\lstset{
backgroundcolor=\color{white},
basicstyle=\fontsize{6.5pt}{6.5pt}\ttfamily\selectfont,
columns=fullflexible,
breaklines=true,
captionpos=b,
commentstyle=\fontsize{6.5pt}{6.5pt}\color{codeblue},
keywordstyle=\fontsize{6.5pt}{6.5pt}\color{codered},
stringstyle=\fontsize{6.5pt}{6.5pt}\color{codepurple},
}
\begin{lstlisting}[language=Python]
# mu: super-resolved images [N, C, H, W]
# hr: high resolution images (ground-truth) [N, C, H, W]
# sigma_pre: predicted standard deviations [N, C, H, W]

sigma_gt = abs(hr - mu) # real residual error
z = randn(hr.shape) # Gaussian random variables (4D tensor)
z = z*(sigma_gt>mean(sigma_gt)) 

# Set z corresponding to small sigma_gt (i.e., easy samples) 
#  to zero and focus on the hard samples.

loss_main = mean(abs(mu+sigma_gt*z.detach()-hr)) # Equation (8)

# As in VAE, averaging the gradient over many samples of z
# during training can handle the expectation over z.

loss_aux = mean(abs(sigma_pre-sigma_gt.detach())) # Equation (10)
return loss_main + beta*loss_aux # beta: penalty factor
\end{lstlisting}
\end{algorithm}

\section{Conclusions}
In this paper, we present a novel learning process for single image super-resolution with neural networks. Instead of best fitting the ground-truth HR images, we introduce a posterior distribution $P(\mathbf{y}|\mathbf{x};\mathbf{W})$ of underlying natural images $\mathbf{y}$. By sampling from $P(\mathbf{y}|\mathbf{x};\mathbf{W})$, the learning process serves as a multiple-valued mapping that better approximates the local manifold corresponding to the observed HR image. The experiments show that our method can produce SR images with minimal artifacts and the predicted standard deviation of $P(\mathbf{y}|\mathbf{x};\mathbf{W})$, as a by-product, allows us to make meaningful estimations on the model uncertainty.

Though presented in the context of super-resolution, the formulated learning process can be extended to other image reconstruction scenarios with $\mathbf{y}$ following different distributions, not limited to Gaussian. Finally, the performance may be further improved by a more complex sigma branch to predict the actual residual error, which helps to solve both HR image regression and uncertainty estimation simultaneously.


{\small
\bibliographystyle{IEEEtran}
\bibliography{egbib}
}

\end{document}